\documentclass[10pt]{article} 
\usepackage[accepted]{tmlr}

\usepackage{amsmath,amsfonts,bm}

\def\eqref#1{equation~\ref{#1}}

\def\1{\bm{1}}

\DeclareMathAlphabet{\mathsfit}{\encodingdefault}{\sfdefault}{m}{sl}
\SetMathAlphabet{\mathsfit}{bold}{\encodingdefault}{\sfdefault}{bx}{n}

\usepackage{booktabs}
\usepackage{hyperref}
\usepackage{url}
\usepackage{comment}
\usepackage{natbib}
\usepackage{graphicx}
\usepackage{caption}
\usepackage{multirow}
\usepackage{algorithm}
\usepackage{subfigure}
\usepackage{tikz}
\usepackage{algpseudocode}
\usetikzlibrary{positioning,arrows.meta, calc, fit}
\usetikzlibrary{decorations.pathreplacing}
\newcommand{\subf}[2]{%
  {\small\begin{tabular}[t]{@{}c@{}}
   \mbox{}\\[-\ht\strutbox]
   #1\\#2
   \end{tabular}}%
}
\def\month{11}
\def\year{2026}
\def\openreview{\url{https://openreview.net/forum?id=8NYTVeeJiF}}

\begin{document}

\title{Deep Adaptive Bayesian Screening} 

%
\author{\name Jade Lejeune Herman \\
\addr Mechatronics, Biostatistics and Sensors (MeBioS) \\
KU Leuven \\
Kasteelpark Arenberg 30, B-3001 Leuven (Heverlee) \\
\email \texttt{jade.lejeuneherman@kuleuven.be}
\AND
\name Arno Strouwen \\
\addr Strouwen Statistics, Hasselt, Belgium \\
Mechatronics, Biostatistics and Sensors (MeBioS) \\
KU Leuven \\
Kasteelpark Arenberg 30, B-3001 Leuven (Heverlee) \\
\email \texttt{contact@arnostrouwen.com} 
\AND
\name Johan A.K. Suykens \\
\addr ESAT-Stadius \\
KU Leuven \\
Kasteelpark Arenberg 10, B-3001 Leuven (Heverlee) \\
\email \texttt{johan.suykens@kuleuven.be} 
\AND
\name Peter Goos \\
\addr Mechatronics, Biostatistics and Sensors (MeBioS) \\
KU Leuven \\
Kasteelpark Arenberg 30, B-3001 Leuven (Heverlee) \\
\email \texttt{peter.goos@kuleuven.be}
}
\maketitle
\begin{abstract}
We introduce Deep Adaptive Bayesian Screening (DABS), a method for performing adaptive factorial screening in discrete design spaces under tight experimental budgets. DABS learns a policy network offline to sequentially select informative experiments, amortizing Bayesian Optimal Experimental Design. It handles binary designs, incorporates sparsity and interactions via a spike-and-slab prior with strong heredity. The model is trained using a contrastive lower bound on information about factor activity with nuisance effect sizes and noise variance analytically integrated out. Unlike prior amortized Bayesian design approaches, DABS also integrates posterior inference at deployment, yielding posterior probabilities of factor activity and credible intervals on effect sizes. We evaluate DABS on simulated screening problems calibrated to
published experimental benchmarks, and show that it matches or outperforms the
strongest non-adaptive Bayesian baseline, with the largest gains when the
budget is small relative to the number of candidate factors.
\end{abstract}
\section{Introduction}
Designing experiments to efficiently explore high-dimensional input spaces is a central challenge in many areas of science and engineering. In particular, screening designs aim to identify the subset of most influential factors among a large number of candidates while minimizing the number of required experimental runs. For instance, in early-stage drug development, researchers must determine which formulation and process variables most strongly affect key quality attributes such as dissolution or stability. When many factors are potentially relevant (say, 12 factors at 2 levels), exploring all possible combinations would require $2^{12}=4096$ experimental runs, which is prohibitively expensive in practice. This combinatorial explosion motivates the use of screening strategies to reduce experimental cost while focusing attention on the variables that drive system performance. 

Classical screening designs are typically non-adaptive and specified a priori (e.g. Plackett-Burman \citep{ref16}, fractional factorial \citep{ref15, ref17} or D-optimal designs \citep{ref23}). Factor significance is then assessed post hoc using frequentist tools such as t-tests or ANOVA. These designs are derived to minimise parameter-estimation variance under a fixed assumed model, rather than to directly target the binary active/inactive decision that screening ultimately requires, and they neither incorporate prior knowledge nor adapt to observed data.

To overcome these limitations, Bayesian optimal experimental design (BOED) \citep{ref5} offers a way to incorporate prior knowledge about factor importance and to select configurations adaptively that maximise the expected information gain about the active factors. 

In the sequential setting, this corresponds to selecting each new experiment to maximise the mutual information between its outcome and the set of active factors, conditioned on the history of past experiments and responses.
However, this requires repeated posterior inference and optimization of an acquisition function (i.e., an information-based utility used to select the next experiment), which is often computationally prohibitive in real-time screening campaigns \citep{ref18}.

Deep Adaptive Design (DAD) \citep{ref1} and related extensions \citep{ref6, RW11} remove this computational burden by training a neural policy network offline on thousands of simulated experimental trajectories. At deployment time, the policy produces the next configuration via a single forward pass, amortizing all optimization. However, these methods target parameter inference rather than active-set recovery and provide no mechanism for posterior inference over factor activity or effect-size credible intervals once the experiments are complete. They are also formulated for continuous design spaces, which requires an additional adaptation step for discrete factorial settings.

To address this gap, we introduce DABS (Deep Adaptive Bayesian Screening). DABS trains a deep policy network entirely offline using simulated trajectories from a discrete design space and complements it with a Gibbs sampler
for posterior inference over active factors and effect sizes. 
DABS assumes a linear--Gaussian response model with a spike-and-slab prior on the effect
sizes and strong heredity on two-factor interactions. These assumptions hold throughout
Sections~3 and~5, and they are what makes the Student-$t$ marginal available in closed form.

The contributions of this paper are threefold. (i) We derive a screening-specific
training objective that targets the mutual information between the experimental
history and the active-set indicators, with the continuous effect sizes and noise
variance marginalised out analytically. This focuses the policy on the screening decision without giving
up the effect sizes, which are recovered at deployment. (ii) We embed screening structure directly into the amortized design
framework via a spike-and-slab prior with strong heredity on two-way
interactions, so that the offline-trained policy produces designs calibrated to
the sparsity and hierarchy of real screening problems. (iii) We couple the amortized policy with a Gibbs sampler at deployment that
exploits the same closed-form Student-$t$ marginal, yielding posterior probabilities of factor activity and credible intervals on
effect sizes from the same runs used to drive the design, with no separate
inference stage, a capability absent from standard policy-based BOED pipelines.
\section{Background}
\subsection{Sequential Bayesian Experimental Design}
Let $\theta$ denote an unknown parameter vector with prior $p(\theta)$. In a sequential experiment, at step $t$, a single new design $\xi_t$ is selected based on the history $h_{t-1} = \{(\xi_s, y_s)\}_{s=1}^{t-1},$
and a scalar response $y_t$ is observed. The informativeness of a candidate design is quantified by the conditional mutual information between $\theta$ and the observation $y_t$ \citep{ref19}:
\begin{equation}
I(\theta;y_t\mid h_{t-1},\xi_t)
=
\mathbb{E}_{\theta\sim p(\theta\mid h_{t-1}),\,y_t\sim p(y_t\mid\theta,\xi_t)}
\left[
\log
\frac{p(y_t\mid\theta,\xi_t)}
     {p(y_t\mid h_{t-1},\xi_t)}
\right].
\end{equation}
The ratio inside the logarithm compares the likelihood under a specific parameter realization $\theta$ with the marginal likelihood averaged over all possible parameter values. 
This quantity corresponds to the expected reduction in posterior entropy about $\theta$ after observing $y_t$. Higher values indicate more informative experiments.

A design policy $\pi_\phi$, parametrized by a neural network with parameters $\phi$, selects experiments adaptively via
$
\xi_t \sim \pi_\phi(\cdot \mid h_{t-1})
$. Following \cite{ref1}, the cumulative expected information gain over an experimental sequence decomposes as 
\begin{equation}
\mathcal{J}(\phi)
=
\mathbb{E}_{\theta \sim p(\theta),h_T \sim p(h_T \mid \theta, \pi_{\phi})}
\left[
\sum_{t=1}^{T}
I(\theta;y_t\mid h_{t-1},\xi_t)
\right].
\label{eq1}
\end{equation}
Maximizing cumulative information gain rather than a single-step criterion encourages globally informative experimental sequences and enables the design policy to account for long-term effects of sequential decisions.
\subsection{Contrastive Estimation of Information Gain}

Direct maximisation of $\mathcal{J}(\phi)$ is intractable because evaluating the marginal likelihood \begin{equation}
p(y_t \mid h_{t-1}, \xi_t) = \int p(y_t \mid \theta, \xi_t)\, p(\theta \mid h_{t-1})\, d\theta
\end{equation}
requires integration over the parameter posterior $p(\theta \mid h_{t-1})$ at every step. \citet{ref1} address this with a Monte Carlo lower bound that replaces the intractable marginal $p(h_T \mid \pi_\phi) = \int p(h_T \mid \theta, \pi_\phi)\, p(\theta)\, d\theta$ by an empirical average over $L$ contrastive samples drawn from the prior, building on earlier stochastic-gradient BOED \citep{ref21}.

Let $\theta_0 \sim p(\theta)$ denote the parameter sample that generates the trajectory $h_T$, and let $\theta_1, \ldots, \theta_L \sim p(\theta)$ denote $L$ independent contrastive samples from the prior. The sequential Prior Contrastive Estimation (sPCE) bound is
\begin{equation}
\text{sPCE}_L(\phi) = \mathbb{E}_{\theta_0 \sim p(\theta),\ h_T \sim p(h_T \mid \theta_0, \pi_\phi),\ \theta_{1:L} \sim p(\theta)}\left[\log \frac{p(h_T \mid \theta_0)}{\frac{1}{L+1}\sum_{i=0}^L p(h_T \mid \theta_i)}\right] \leq \mathcal{J}(\phi),
\label{eq4}
\end{equation}
where the expectation is over $\theta_0 \sim p(\theta)$, the trajectory $h_T \sim p(h_T \mid \theta_0, \pi_\phi)$, and the contrastive draws $\theta_{1:L} \sim p(\theta)$. A corresponding upper bound yields the sandwich relation \citep{ref20}
\begin{equation}
\text{sPCE}_L(\phi) \leq \mathcal{J}(\phi) \leq \text{sNMC}_L(\phi),
\label{eq5}
\end{equation}
with both bounds converging to $\mathcal{J}(\phi)$ as $L \to \infty$. The sPCE bound serves as the training objective, and the corresponding sNMC
upper bound can be used at evaluation time to verify that the sPCE bound is
tight.
\section{Deep Adaptive Bayesian Screening}
\subsection{Problem Setup and Factorial Screening Model}
Consider a screening experiment with $p$ factors, each set to a high or low level at every experimental run \citep{ref15}.
The objective is to identify the active set from as few experiments as possible, together with credible intervals on the magnitudes of the corresponding effects. At step $t=1,\ldots,T$, a design vector $\xi_t = (\xi_{t,1},...,\xi_{t,p}) \in \{-1,+1\}^p$ is selected where $\xi_{t,k} = +1$ (resp. $-1$) indicates that factor $k$ is set to its high (resp. low) level and a scalar response is observed:
\begin{equation}
y_t = f(\xi_t,\theta) + \varepsilon_t,
\quad
\varepsilon_t \sim \mathcal{N}(0,\sigma^2), \quad\sigma^2 \sim \mathrm{Inv\text{-}Gamma}(\alpha_0,\beta_0), \quad \alpha_0, \beta_0 > 0,
\label{eq:model}
\end{equation}
with the conjugate inverse-gamma prior on $\sigma^2$ enabling the analytical marginalization described in Section 3.2. The response function $f$ is linear in the effect sizes, as specified in Eq.~\ref{eq9bb}, so
the model is linear--Gaussian throughout. The parameter vector $\theta = (z, \beta, \sigma^2)$ combines binary activity indicators $z \in \{0,1\}^p$, continuous effect sizes $\beta \in \mathbb{R}^p$ and the noise variance. An indicator $z_k = 1$ (resp. $0$) denotes that factor $k$ is active (resp. inactive) and $\beta_k$ determines the magnitude of its contribution. We restrict attention to binary designs $\xi_t \in \{-1,+1\}^p$, which is the standard
convention in factorial screening, where each factor is set to a high or a low level
\citep{ref15}. For models that are linear in the inputs $\xi$, this restriction is
also supported theoretically. 

\textbf{Proposition 1.} \emph{Consider the linear--Gaussian model $y = \xi^\top \beta + \varepsilon$ with
$\varepsilon \sim \mathcal{N}(0, \sigma^2)$ and $\beta \sim \mathcal{N}(0, \Sigma_\beta)$,
over the design space $\xi \in [-1,1]^p$. For information-theoretic design criteria that
depend on $\xi$ through a quadratic form $\xi^\top \Sigma \xi$ with $\Sigma \succeq 0$,
there exists an optimal design in $\{-1,+1\}^p$.} A proof is provided in Appendix~\ref{apAa}.

Proposition 1 concerns the classical single-step criterion at a fixed active set.
The sPCE objective of Section~3.2 averages over the active set and over the full horizon of
$T$ experiments, so it does not have the quadratic-form structure the proposition requires.
We therefore treat the proposition as motivating the design space rather than as
establishing optimality of binary levels for the sPCE objective itself.

Screening experiments rely on the factor sparsity principle \citep{ref7}, which posits that generally only a small subset of factors significantly influence the response. This assumption is encoded through a spike-and-slab prior on the effect sizes, which is a standard approach for Bayesian variable selection  \citep{ref4}:
\begin{equation}
\rho \sim \mathrm{Beta}(a, b), \qquad
z_k \mid \rho \sim \mathrm{Bernoulli}(\rho), \qquad
\beta_k \mid z_k,\sigma^2 \sim
\begin{cases}
\mathcal{N}(0, \sigma^2 \ \tau_{\text{active}}^2), & z_k = 1, \\
\mathcal{N}(0, \sigma^2 \tau_{\text{inactive}}^2), & z_k = 0 .
\end{cases}
\label{eq8}
\end{equation}
where $\rho \in (0, 1)$ is the prior probability that a factor is active, drawn from a $\mathrm{Beta}(a, b)$ hyperprior with prior mean $\mathbb{E}[\rho] = a / (a + b)$, $\tau^2_{\text{active}}$ is the variance for active effects, and $\tau^2_{\text{inactive}} \ll \tau^2_{\text{active}}$ shrinks inactive effects toward zero.
Besides main effects, pairwise interactions between factors, which capture joint effects beyond individual contributions, are also considered. Each interaction between factors $i$ and $j$ is associated with an effect size $\beta_{ij}$ and an activity indicator $z_{ij}$. To reflect the structure typically assumed in screening experiments, we impose the strong heredity principle \citep{ref14, ref24, ref25}, meaning that an interaction between factors $i$ and $j$ can only be active if both parent main effects are active. The hyperparameter $\pi_{\mathrm{int}} \in (0, 1]$ is the conditional probability that a heredity-allowed interaction is active:
\begin{equation}
z_{ij} \mid z_i, z_j \sim \mathrm{Bernoulli}(\pi_{\mathrm{int}} \cdot z_i z_j),
\qquad
\beta_{ij}\mid z_{ij}, \sigma^2\sim
\begin{cases}
\mathcal{N}(0,\sigma^2 \tau_{\text{active}}^2), & z_{ij} = 1, \\
\mathcal{N}(0,\sigma^2 \tau_{\text{inactive}}^2), & z_{ij} = 0 .\end{cases}
\label{eq9}
\end{equation}
Strong heredity is a modelling assumption. Appendix \ref{apF} reports an ablation that relaxes it and quantifies the sensitivity of the recovered active set. The expected response is modeled as a linear combination of all effects:
\begin{equation}
f(\xi_t,\theta)
=
\sum_{k=1}^{p} \beta_k\xi_{t,k}
+
\sum_{i<j} \beta_{ij}\xi_{t,i}\xi_{t,j}.
\label{eq9bb}
\end{equation}The activity indicators enter through the prior variances of Eqs.~\ref{eq8}
and~\ref{eq9} rather than through the response. The formulation captures two key properties of screening experiments: the sparsity of active factors and structured interactions through the heredity constraint. Under
heredity, the space of admissible activity patterns reduces from
$\{0,1\}^{p+p(p-1)/2}$ to those patterns in which $z_{ij}=1$ only if
$z_i = z_j = 1$. For instance, when $p = 15$, this is a reduction from $2^{120}$ to about $2^{105}$ distinct configurations, a reduction of more than four orders of magnitude. Because the spike-and-slab prior is sparse, few main effects are active in a typical draw, so only a small number of interaction indicators are free. This reduction, induced by the strong
heredity assumption, is critical for scalability as it renders the contrastive sPCE
objective tractable: contrastive samples are drawn hierarchically, first
the main-effect indicators and then only 
from the interaction indicators permitted by the heredity constraints, rather than from the full joint space.
\subsection{Analytical Marginalisation} 
The screening model of Section 3.1 decomposes the parameter 
vector as $\theta = (z, \beta, \sigma^2)$, where $z = ((z_k)_{k=1}^p, (z_{ij})_{i<j})$ denotes the joint activity indicators, $\beta$ the continuous effect sizes for main effects and heredity-constrained interactions, and $\sigma^2$ the observation noise
variance. Under the Gaussian likelihood of Eq. \ref{eq:model} and the
conjugate normal--inverse-gamma prior on $(\beta, \sigma^2)$ induced by
the spike-and-slab specification in Eq. \ref{eq8} together with the inverse-gamma prior on $\sigma^2$, the continuous parameters admit
closed-form integration. Conditioned on $z$, the joint distribution of the
history $h_T = \{(\xi_t, y_t)\}_{t=1}^{T}$ follows a multivariate 
Student-$t$ distribution
\begin{equation}
p(h_T \mid z) \;=\; \iint p(h_T \mid z, \beta, \sigma^2)\,
p(\beta \mid z, \sigma^2)\, p(\sigma^2)\, d\beta\, d\sigma^2,
\label{eq:student-marginal}
\end{equation}
whose scale matrix and degrees of freedom are determined by the active set
$z$, the design matrix induced by the history, and the prior hyperparameters
$(\alpha_0, \beta_0, \tau^{2}_{\mathrm{active}}, \tau^{2}_{\mathrm{inactive}}, \pi_{\mathrm{int}})$.
The full derivation is provided in Appendix~\ref{apAc}.

This analytical marginalization has two consequences for training. First, the sPCE objective (Eq.~\ref{eq4}) specializes naturally: contrastive samples need only be drawn from $p(z)$ over the discrete activity indicators, with each trajectory likelihood $p(h_T \mid z_i)$ evaluated via the closed-form Student-$t$ marginal. Second, because $\beta$ and $\sigma^2$ never appear in the contrastive draws, the Monte Carlo variance of the estimator does not inflate with the dimension of the continuous parameters, keeping the training signal stable as $p$ increases.
The resulting estimator therefore targets the cumulative mutual information $I(z; h_T \mid \pi_\phi)$ between the activity indicators and the trajectory under the policy, rather than the full $I(\theta; h_T \mid \pi_\phi) = \mathcal{J}(\phi)$ used in DAD-style methods. This concentrates the training signal on the discrete screening decision. The continuous parameters $(\beta, \sigma^2)$ are integrated out analytically during
training and recovered at deployment by a Gibbs sampler whose conditionals for
$(\beta, \sigma^2)$ are available in closed form. 
The same Student-$t$ marginal of Eq.~\ref{eq:student-marginal} is used at both
stages, in the contrastive objective during training and in the activity update at
deployment.
\subsection{Learning the Adaptive Design Policy}
The conjugate linear–Gaussian structure yields a closed-form marginal likelihood, eliminating the need for variational bounds or nested Monte Carlo estimators typically required in Bayesian optimal experimental design. As a result, policy learning reduces to direct maximization of the sPCE objective using standard gradient-based methods. 

However, the designs $\xi_t \in \{-1,+1\}^p$ are discrete. Therefore, pathwise gradient estimators cannot be applied directly. To enable end-to-end training while preserving discrete designs at evaluation, we employ a straight-through Gumbel–Softmax relaxation \citep{ref2, ref10}. At each step, the policy network outputs logits $\ell_t = \pi_\phi(h_{t-1}) \in \mathbb{R}^p$, and a design is sampled as \begin{equation}
   \xi_t = \mathrm{sign}(\ell_t + g_t), \quad g_t \sim \mathrm{Gumbel}(0, \tau), \label{eq11}
\end{equation}with temperature $\tau$ annealed from $\tau_{\text{start}}$ to $\tau_{\text{end}}$ during training. The forward pass uses the discrete $\{-1,+1\}$ designs for likelihood evaluation, while the backward pass propagates gradients through the continuous relaxation \citep{ref27}. At evaluation time, Gumbel noise is removed and logits are thresholded at zero, yielding deterministic $\{-1,+1\}$ designs.

Because the history $h_{t-1}$ grows with $t$, we compress it into fixed-dimensional
sufficient statistics before passing it to the policy. Specifically, we map the statistics
$(X_{t-1}^\top X_{t-1}, X_{t-1}^\top y_{t-1}, y_{t-1}^\top y_{t-1})$, formed from the
expanded design matrix $X_{t-1}$ defined in Section~3.4, through a two-hidden-layer
multilayer perceptron (MLP) that outputs the logits $\ell_t$. These statistics fully
determine the Student-$t$ posterior under the conjugate prior. This representation is exactly sufficient for Bayesian inference under the
linear--Gaussian likelihood, and its dimension is constant in $T$. Appendix~\ref{apA} compares it against a set-equivariant encoder applied
directly to the raw $(\xi_s, y_s)$ pairs. The set-equivariant encoder is the
stronger of the two on the ablation problem class of Appendix~\ref{apF}. 
The encoder architecture and a formal sufficiency argument are given
in Appendix~\ref{apA}.

A practical failure mode in sequential design policies is collapse, where the policy repeatedly selects identical or highly similar designs across steps, effectively ignoring the adaptive nature of the task \citep{b4}. In screening, this is particularly problematic: repeating the same design provides no additional information, as observations are conditionally independent given $\theta$. Empirically, optimizing sPCE alone does not prevent collapse, since increasing logit magnitudes drive the softmax toward a degenerate deterministic policy long before exploration has resolved the posterior over $z$. 

To mitigate this, we add a squared-$\ell_2$ penalty on the policy logits \citep{rw23, rw24}:
$ \mathcal{L}_{L2}(\phi) = \frac{1}{T} \sum_{t=1}^{T} \|\ell_t\|_2^2,$ which keeps logits bounded and preserves Gumbel-induced exploration throughout training. The full objective is\begin{equation}
\mathcal{L}(\phi) = -\widehat{\text{sPCE}}_L(\phi) + \mu\, \mathcal{L}_{\text{L2}}(\phi),
\end{equation}where $\mu$ controls the trade-off between information gain and policy regularization. We minimize $\mathcal{L}(\phi)$ via stochastic gradient descent, which is equivalent to maximizing the sPCE lower bound subject to an $\ell_2$ penalty on the policy logits. Adding the penalty does shift the stationary points of the combined objective, but its strength grows with the logit magnitude, so it acts primarily in the runaway-logit regime that leads to collapse. Because identical designs carry no additional information under conditional independence, a policy that maximizes $\mathcal{J}(\phi)$ naturally produces diverse designs with moderate logit magnitudes, for which the penalty is negligible. Its role is therefore to prevent the large-logit collapse to which Gumbel-Softmax training is prone, rather than to trade off information gain. The full training procedure is summarized in the pseudo code in Appendix \ref{apAd}.
\subsection{Policy Network Architecture}
Having defined the policy optimization procedure, we now describe the parametrization of the design policy $\pi_\phi(\xi_t \mid h_{t-1})$. The policy must map a variable-length history of experiment-outcome pairs to the next design, with two structural requirements: (i) invariance to the order of past experiments and (ii) accommodation of histories of growing length.

In the screening setting, observations are conditionally independent given the true parameters, $y_t \perp h_{t-1} \mid \theta, \xi_t$. As shown by \citet{ref1}, this implies that the optimal policy depends only on the multiset of past $(\xi_s, y_s)$ pairs, not their order. We exploit this structure by basing $\pi_\phi$ on the sufficient statistics of the linear–Gaussian likelihood, which are permutation-invariant by construction.

Given history $h_{t-1}$, let $X_{t-1} \in \mathbb{R}^{(t-1) \times d}$ be the
expanded design matrix whose $s$-th row is
$(\xi_{s,1}, \ldots, \xi_{s,p}, \xi_{s,1}\xi_{s,2}, \ldots, \xi_{s,p-1}\xi_{s,p})$,
comprising the $p$ main-effect columns followed by the $p(p-1)/2$ two-factor interaction
columns, and let $y_{t-1} \in \mathbb{R}^{t-1}$ collect the corresponding responses. We compute
\begin{equation}
    s_{t-1} = \left( \mathrm{vec}\!\left(\frac{X_{t-1}^\top X_{t-1}}{t-1}\right),\;
\frac{X_{t-1}^\top y_{t-1}}{t-1},\; \frac{y_{t-1}^\top y_{t-1}}{t-1} \right)
\in \mathbb{R}^{\frac{d(d+1)}{2} + d + 1}
\label{eq:14}
\end{equation}
where $d=p+p(p-1)/2$ and $\mathrm{vec}(\cdot)$ extracts the upper-triangular entries of the symmetric matrix $X_{t-1}^\top X_{t-1}$. They arise by completing the square in the Gaussian likelihood, which depends on the history
only through $X_{t-1}^\top X_{t-1}$, $X_{t-1}^\top y_{t-1}$ and $y_{t-1}^\top y_{t-1}$. These statistics are exactly sufficient for the Student-$t$ posterior over $(\beta, \sigma^2)$ under the conjugate Normal–Inverse-Gamma prior (Section 3.2), and their dimension is constant in $T$. 

The policy logits are produced by a single MLP: 
    $\ell_t=\mu_{\phi}(s_{t-1},t) \in \mathbb{R}^p,$
where $t$ is a scalar positional index appended to the input. The MLP $\mu_\phi$ has two hidden layers with Softplus activations and applies dropout during training to prevent logit saturation. From $\ell_t$, the design $\xi_t \in \{-1,+1\}^p$ is produced via the straight-through Gumbel-Softmax mechanism described in Section 3.3.

This design has two key properties. Permutation invariance over past experiments holds by construction, since the statistics in Eq. \ref{eq:14} are symmetric functions of the $(\xi_s, y_s)$ pairs. Moreover, $\pi_\phi$ is the only trainable component of the system: leveraging the closed-form Student-$t$ marginal, the contrastive objective (Eq. \ref{eq4}) can be evaluated analytically, without requiring a separate utility or critic network. As a result, training reduces to backpropagating a single likelihood-based loss through a single MLP, avoiding the instabilities often associated with multi-network adversarial training. Figure~\ref{fig:init}(a) provides a visual overview.
\subsection{Posterior Inference and Deployment}
Following the completion of $T$ adaptive experiments, the collected history $h_T$ is used to identify the active factors and estimate their effects via the posterior distribution $p(z, \beta \mid h_T)$. Unlike prior policy-based BOED methods, which focus solely on design, DABS integrates posterior inference into deployment, enabling both variable selection and posterior uncertainty over the active set.

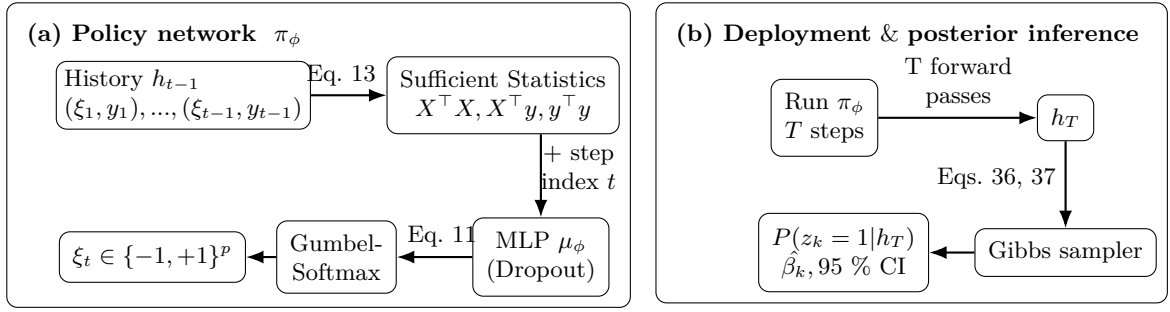
\begin{figure}[!h]
\centering
\begin{tikzpicture}[
    font=\small,
    node distance=8mm,
    box/.style={draw, rounded corners, align=center, inner sep=5pt},
    arrow/.style={-Latex, thick}
]
\node[align=center] (titlea) at (-4.5,0) 
{\textbf{(a) Policy network } $\pi_\phi$ };
\node (hist) [box, minimum width=1.7cm,
            minimum height=2mm, inner sep= 3pt, align=left, xshift=-18mm] at(-2.5,-0.8) {History $h_{t-1}$ \\$(\xi_1, y_1),...,(\xi_{t-1}, y_{t-1})$};
\node[box, right=10mm of hist] (stat) {Sufficient Statistics \\ $X^\top X, X^\top y, y^\top y$};
\node[box, below=11mm of stat, xshift=5mm] (emitter) {
MLP $\mu_\phi$ \\
(Dropout)
};
\node[box, left=10mm of emitter] (gumbel) {
Gumbel- \\ Softmax
};
\node[box, left=4mm of gumbel] (xi) {$\xi_t \in \{-1, +1 \}^p$};
\draw[arrow] (hist)-- node[pos=0.6, above, xshift=2 mm, xshift=-4mm] {Eq.~\ref{eq:14}} (stat);
\draw[arrow] (stat.south)++(0.5, 0) -- 
node[pos=0.6, above, xshift=5 mm, yshift=-2mm, align=center, text width=2.5cm] 
{$+$ step\\index $t$} 
(emitter.north);
\draw[arrow] (emitter) -- node[pos=0.6, above, xshift=2 mm] {Eq.~\ref{eq11}} (gumbel);
\draw[arrow] (gumbel) -- (xi);
\node[align=center, right=4.6cm of titlea] (titleb)
{\textbf{(b) Deployment} \& \textbf{posterior inference}};
\node[box, below=3mm of titleb, xshift=-11mm] (run) {Run $\pi_\phi$ \\ $T \ \text{steps}$};
\node[box, right=21mm of run] (ht) {$h_T$};
\node[box, below=11.5mm of ht] (gibbs) {Gibbs sampler};
\node[box, left=6mm of gibbs] (pzk) {$P(z_k=1|h_T)$ \\ $\hat{\beta_k}, \text{95 \% CI}$};
\draw[arrow] (run) -- 
node[pos=0.6, above, xshift=-2 mm, align=center, text width=2.5cm] 
{T forward\\passes} 
(ht);
\draw[arrow] (ht) -- node[pos=0.6, left, yshift = 2 mm] {Eqs.~\ref{20},~\ref{21}} (gibbs);
\draw[arrow] (gibbs) -- (pzk);
\coordinate (topref) at ($(titleb.north)$);
\coordinate (topref2) at ($(titlea.north)$);
\coordinate (botref) at ($(emitter.south)$);
\coordinate (botref2) at ($(gibbs.south)$);
\node[
    draw,
    rounded corners,
    inner sep=4pt,
    fit=(topref2)(botref)(titlea)(stat)(xi)(hist)(emitter)
] (blockA) {};
\node[
    draw,
    rounded corners,
    inner sep=4pt,
    fit=(topref)(botref2)(titleb)(run)(ht)(gibbs)(pzk)
] (blockB) {};
\end{tikzpicture}
\setlength{\abovecaptionskip}{3pt}
\caption{(a) DABS policy network $\pi_{\phi}$ and (b) deployment with Gibbs posterior inference.}
\label{fig:init}
\end{figure}

Direct posterior computation is intractable due to the discrete nature of $z \in \{0,1\}^p$. 
We therefore employ a Gibbs sampler \citep{ref8, ref12, ref30} that alternates
between sampling $(\beta, \sigma^2) \mid z, h_T$ from their closed-form
conditionals and updating the activity indicators from the Student-$t$ marginal
of Eq.~10. Strong heredity makes the admissible set of indicators
non-rectangular, so each main-effect indicator is updated jointly with the
interactions it appears in, through a Metropolis-Hastings move whose acceptance
ratio is available in closed form (see Algorithm 2 in Appendix~\ref{apB} for the
full deployment pipeline, and Appendix~\ref{apC} for the step-by-step details).
The primary outputs are the posterior activity probabilities and effect size estimates:\begin{equation}
\hat{P}(z_k = 1 \mid h_T) = \frac{1}{G} \sum_{g=1}^{G} z_k^{(g)}, 
\qquad
\hat{\beta}_k = \frac{1}{G} \sum_{g=1}^{G} \beta_k^{(g)}.
\end{equation} with analogous expressions for the interaction indicators $z_{ij}$ and effects $\beta_{ij}$. A factor (or interaction) is declared active if $\hat{P}(z_k = 1 \mid h_T) > 0.5$,
and uncertainty is quantified via posterior credible intervals, whose calibration
is evaluated in Appendix~\ref{apI}. At the 95\% level these intervals are slightly below nominal for main effects
whose activity is correctly identified, reaching between $0.916$ and $0.934$
across the three case studies, and are lower for active interactions at
approximately $0.75$.
The full inference
procedure is summarized in Figure~\ref{fig:init}(b).
\setlength{\textfloatsep}{6pt}

\subsection{When Does Adaptivity Help }
The screening model of Sections 3.1--3.2 decomposes $\theta = (z, \beta, \sigma^2)$ into discrete activity indicators, continuous effect sizes, and the noise variance. This decomposition determines which component of $\theta$ carries the adaptive gain and which does not. By the chain rule of mutual information, the total information the experimental history $h_T$ carries about $\theta$ factorises as
\begin{equation}
I(\theta; h_T) = I(z; h_T) + \mathbb{E}_{z}\!\left[I(\beta, \sigma^2 ; h_T \mid z)\right].
\label{eq:chain-rule}
\end{equation}
The first term measures information about the discrete active-set indicators. The second measures information about the effect sizes and noise variance conditional on the active set. The sPCE objective used by DABS (Eq.~\ref{eq4}) targets the first term via the analytical Student-$t$ marginalisation of Section~3.2, integrating out $\beta$ and $\sigma^2$.  \\
The conditional term does not provide an additional source of adaptive gain. Indeed, decomposing it once more as
$I(\beta, \sigma^2; h_T \mid z) = I(\sigma^2; h_T \mid z) + I(\beta; h_T \mid z, \sigma^2)$,
both parts are determined by the design alone. Given $z$ and $\sigma^2$, the model is
linear--Gaussian with known variance, and the classical result on Bayesian experimental
design in this setting \citep{ref19,ref5} gives
$I(\beta; h_T \mid z, \sigma^2) = \tfrac{1}{2}\log\det\!\big(I + \mathrm{diag}(\tau_z^2)\, X^\top X\big)$,
which depends on the design matrix but neither on the observed responses nor on $\sigma^2$.
Marginalising over $\beta$ gives $y \mid \sigma^2, z \sim \mathcal{N}(0, \sigma^2 \Sigma_z)$ with
$\Sigma_z = I_T + X \mathrm{diag}(\tau_z^2) X^\top$. Hence, the sufficient statistic
$y^\top \Sigma_z^{-1} y$ is distributed as $\sigma^2 \chi^2_T$ irrespective of the design. It follows that the
information carried by the history about $\sigma^2$ depends only on the budget $T$. \\ \\
\begin{minipage}{0.35\textwidth}
\centering
\begin{tikzpicture}[scale=0.15, font=\small]
  \fill[blue!15] (0,0) -- (28,0) -- (28,28) -- cycle;
  \fill[blue!30] (0,0) -- (28,0) -- (28,14) -- cycle;
  \draw[->] (0,0) -- (30,0) node[right] {$p$};
  \draw[->] (0,0) -- (0,28) node[above] {$T$};
  \draw[very thick] (0,0) -- (28,28) node[right] {$T = p$};
  \node[align=center, font=\footnotesize] at (7, 22) {static $\approx$ DABS \\ (identifiable)};
  \node[align=center, font=\footnotesize] at (22, 6) {DABS \\ advantage};
  \draw[->, thick, red!80!black] (15, 4) -- (15, 17);
  \node[red!80!black, right, font=\scriptsize] at (15.5, 11) {\S5.2};
  \draw[->, thick, blue!80!black] (10, 16) -- (26, 16);
  \node[blue!80!black, above, font=\scriptsize] at (21, 16.5) {\S5.3};
  \draw[->, thick, orange!80!black] (6, 6) -- (6, 25);
  \node[orange!80!black, right, font=\scriptsize] at (6.5, 15) {\S5.4};
  \foreach \x in {5, 10, 15, 20, 25}
    \draw (\x, 0) -- (\x, -0.4) node[below, font=\scriptsize] {\x};
  \foreach \y in {5, 10, 15, 20, 25}
    \draw (0, \y) -- (-0.4, \y) node[left, font=\scriptsize] {\y};
\end{tikzpicture}
\captionsetup{width=.9\linewidth}

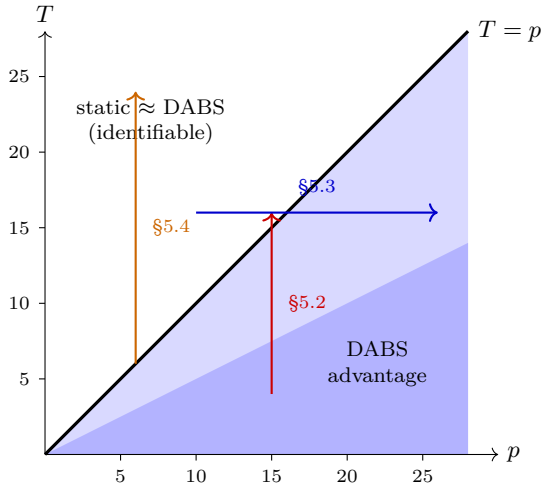
\captionof{figure}{The wedge prediction: below the boundary $T = p$ for main-effect
screening, DABS opens an advantage over the optimal static design, with arrows marking
each case study's sweep.}
\label{fig:wedge}
\end{minipage}
\begin{minipage}{0.65\textwidth}
Since neither term depends on the realised responses, an optimal non-adaptive design achieves the same conditional information gain as any adaptive policy at equal budget, and all adaptive leverage in Eq.~\ref{eq:chain-rule} flows through $I(z; h_T)$. We define the identifiability boundary as the budget at which $T$ equals the number of
terms being screened, namely $p$ for main-effect screening and $p + \binom{p}{2}$ when
two-factor interactions are included. This is a practical reference point rather than a
sharp information-theoretic threshold, since sparsity can allow the active set to be
resolved from fewer runs. Above this boundary the design has enough capacity to span the
coordinate directions required to resolve $z$, and no fixed-versus-adaptive gap remains. Below this identifiability boundary, DABS can use early observations to identify candidate actives and steer subsequent queries into the relevant subspace, while a static policy must commit to a $T$-dimensional measurement subspace before any data are observed. This predicts a wedge in the $(p, T)$ plane, illustrated in Figure~\ref{fig:wedge}. The experiments of Section~5 test this prediction along complementary axes: sweeping $T$ at fixed $p$ (Section~5.2), sweeping $p$ at fixed $T$ (Section~5.3), and sweeping $T$ at three problem sizes under strong-heredity interaction recovery (Section~5.4).
\end{minipage}
\section{Related work}
Screening designs are a core component of design of experiments for efficiently identifying active factors with few runs. Traditional methods \citep{RW3} (fractional factorial \citep{ref17}, Plackett-Burman \citep{ref16} and supersaturated designs \citep{RW2}) focus on estimating main effects and, when resolution permits, interactions under sparsity assumptions. These methods are non-adaptive by construction: the experimental plan is fixed before any observations are collected. When the goal is to identify active factors under a tight budget, this static structure can be inefficient, as it allocates runs uniformly without exploiting the sparsity of the active set \citep{b5}, and ignores information that early observations provide about uninformative directions. 
Bayesian screening methods instead quantify uncertainty over the active-factor set and model sparsity explicitly through priors such as spike-and-slab \citep{RW6,ref30} or horseshoe \citep{RW7}. These methods focus on posterior inference and variable selection but typically assume a fixed experimental design. Recent work extends these ideas to ultra-high-dimensional regression \citep{RW8} or double-robust variable selection \citep{RW9}. The foundational framework for sequential experiment selection is provided by Bayesian Optimal Experimental Design formalized by \cite{ref5}, which selects experiments by maximizing an information theoretic utility.  However, in practice, these methods are not amortized: each new experiment requires an expensive optimization step at decision time \citep{RW13,RW17,RW18}, limiting use in sequential settings. A widely used simplification of classical BOED is greedy one-step lookahead \citep{RW32,RW33}, which we use as the Myopic baseline in Section~5.1.

Amortized policy-based Bayesian experimental design was introduced with Deep Adaptive Design (DAD) \citep{ref1} and its likelihood-free extension iDAD \citep{ref6} (we do not adopt this variant, see Appendix \ref{apD}), which learn a policy offline to map past observations to new experimental designs. This avoids the expensive per-step optimization required by classical Bayesian optimal experimental design \citep{ref5}. Subsequent work extends this framework to continuous dynamical systems \citep{RW22}. These methods optimize cumulative information gain during training. Posterior inference at deployment is typically delegated to a separate mechanism \citep{RW21}. Recent work has explored targeted experimental design objectives that focus on specific functionals of the parameters rather than total information (e.g., contextual optimization in CO-BED \citep{RW31}, predictive information gain in EPIG \citep{RW30}). In contrast, we target information about the support indicators of active factors. However, existing amortized approaches were developed for continuous-parameter estimation and do not incorporate the sparsity or heredity structure of screening problems. Adapting them to the discrete factorial regime with binary designs
$\{-1, +1\}$, strong-heredity interactions, and budgets that are tight relative
to the number of terms being screened is the gap we address.
\section{Results}
\subsection{Experimental Setup}
We evaluate DABS on three published screening benchmarks calibrated to published experimental settings and compare it against four baselines spanning Bayesian design, classical DoE, and adaptive sensing. The benchmarks are chosen to test the wedge prediction of Section~3 across complementary directions. We first describe the evaluation protocol before introducing the case studies and baselines. Every experiment reports F1, true positive rate (TPR) and false discovery rate (FDR) computed from the Gibbs decoder at posterior inclusion threshold 0.5. Metrics are reported separately for main effects and two-factor interactions. Error bars denote $95\%$ confidence intervals over five independent random seeds for each $(p,T)$ setting.

Section~3 predicts that adaptivity helps only when the budget $T$ is insufficient to identify the active set $z$, since under the linear-Gaussian model adaptation cannot improve estimation of effect sizes conditional on $z$ \citep{ref19,ref5}. Once this identifiability boundary is crossed, adaptive and optimal fixed designs perform similarly. The experiments below test this by varying either the budget or the problem size.

We consider three published screening benchmarks. For each benchmark, we simulate responses under the model of Section~3 with hyperparameters calibrated to the published problem. Case Study 1 uses the biotechnological benchmark of inulinase production by \textit{A. niger} \citep{s2}, with $p = 15$ candidate factors, calibrated so
that three main effects are active in expectation, while varying $T$ from 4 to 16. Case Study 2 uses the biomedical benchmark of hematopoietic stem-cell expansion in cord blood \citep{s3}, fixing $T=16$ and increasing $p$ from 10 to 26, causing the expected number of active factors to grow from roughly four to ten. Case Study 3 uses the metallurgical benchmark of electrochemical dissolution of
GH4738 nickel-based superalloy scrap \citep{s6}, whose published analysis
identifies four active main effects and three two-factor interactions under
strong heredity. Here we vary $T$ for $p \in \{4, 6, 8\}$.

At evaluation the ground truth is drawn from the generative model of
Section~3.1. For each instance we draw $\sigma^2 \sim \mathrm{InvGamma}(\alpha_0,
\beta_0)$, the activity indicators, and finally the effect sizes from Eqs.\ref{eq8}
and \ref{eq9} conditional on $\sigma^2$. Because the slab is scaled by $\sigma$, the
standardised effect of an active factor satisfies
$\beta_k / \sigma \sim \mathcal{N}(0, \tau_{\text{active}}^2)$. We therefore report the signal-to-noise ratio as $\tau_{\text{active}}^2$, which is $4$, $1$ and $4$ for Case Studies 1, 2 and 3 respectively, and does not
depend on $\sigma$.

We compare DABS with four baselines. The \emph{static design} is the optimal fixed design under the same Bayesian model and sPCE objective, isolating the benefit of adaptivity. The \emph{Plackett--Burman} design \citep{ref16} provides the standard orthogonal fractional-factorial baseline. The \emph{myopic} baseline greedily maximises one-step expected information gain under the same Bayesian model, isolating the value of the learned policy. Finally, the \emph{multi-stage adaptive} baseline from compressed sensing \citep{pw2} iteratively prunes candidate factors using marginal-effect estimates. The static and myopic methods therefore serve as matched Bayesian controls, while the latter two compare DABS against established approaches from neighbouring literatures. Within each case study, every method is evaluated under identical generative
settings, drawn from the same prior with the same hyperparameters. Code, model checkpoints and evaluation scripts are available at \url{https://github.com/JadeLej/DABS_code}.
\subsection{Main effects screening (biotechnology benchmark)}
This first case study considers the biotechnological benchmark of inulinase production by A.~niger \citep{s2}. The experiment involves $p = 15$ controllable factors: four carbon sources (glucose $X_1$, fructose $X_2$, inulin $X_3$, sucrose $X_4$), eight nitrogen sources ($X_5, \ldots, X_{12}$, including yeast extract $X_5$ and peptone $X_6$), medium pH ($X_{13}$), inoculum level ($X_{14}$), and trace element concentration ($X_{15}$). The published analysis identifies $X_3$, $X_5$, and $X_{13}$ as the three factors with a statistically significant effect on inulinase activity.

We simulate data from the spike-and-slab model of Section~3 with $\rho$ fixed at
$0.2$, $\tau_{\text{active}} = 2.0$, $\tau_{\text{inactive}} = 0.01$,
$\alpha_0 = 3.0$ and $\beta_0 = 1.0$. Interaction indicators are drawn with $\pi_{\text{int}} = 0.5$ under strong heredity, as in the other two case studies. This case study reports main-effect metrics only, and interaction recovery is treated in Section 5.4.
The budget is swept over $T \in \{4, 6, 8, 10, 12, 14, 16\}$, spanning the under-identified regime $T < p$ and reaching the identifiability boundary $T = p = 15$. For each $(T, \text{seed})$ cell, we train one DABS policy, evaluated over 100 rollouts on main-effect F1 and TPR. FDR curves are reported in Appendix~\ref{app:5_1_extra}.
\begin{figure}[!h]
\centering
\begin{tabular}{cc}
\subf{\includegraphics[width=78mm]{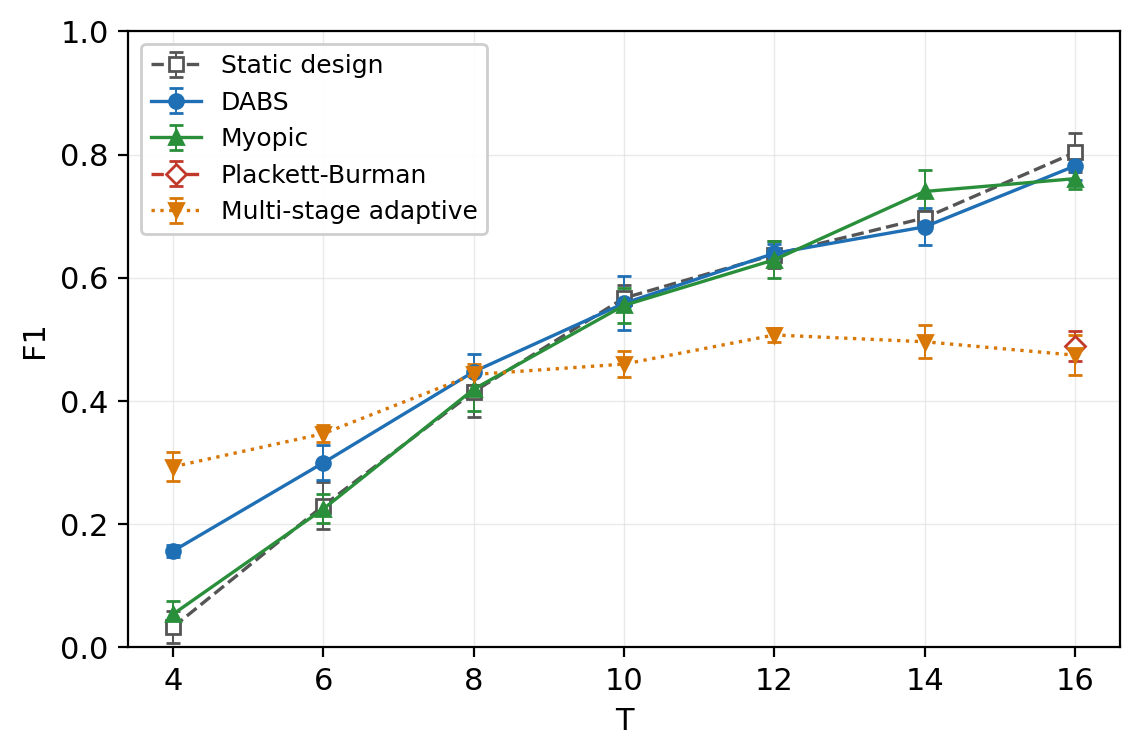}}
      {} &
\subf{\includegraphics[width=78mm]{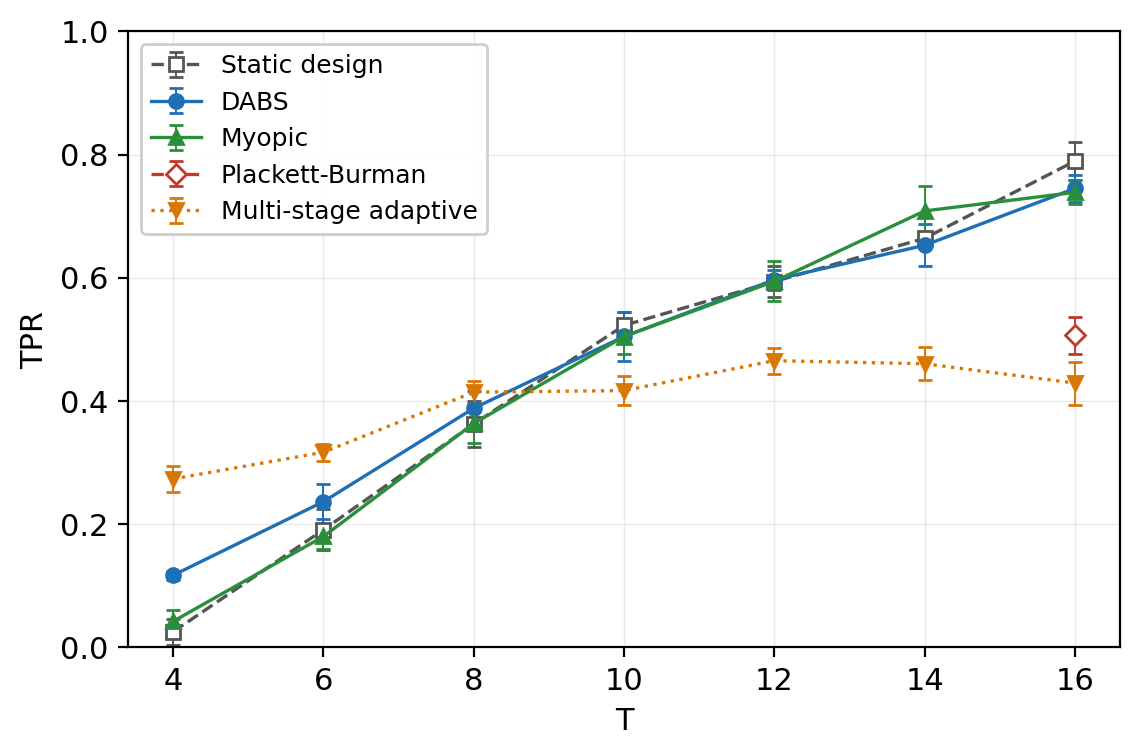}}{}
\end{tabular}
\caption{Main-effect F1 (left) and TPR (right) versus budget $T$ at $p = 15$. Bands are 95\% CIs over 5 seeds.}
\label{fig:c1_metrics}
\end{figure}

Figure~\ref{fig:c1_metrics} shows main-effect F1 (left) and TPR (right) versus
the budget $T$. In the deep-screening regime at $T \leq 6$, DABS and the
multi-stage adaptive baseline separate from the static design and the myopic
baseline. At $T = 4$, static F1 is $0.03$ and the myopic baseline $0.05$,
against $0.16$ for DABS and $0.29$ for multi-stage adaptive. Three of five
static seeds collapse to F1 below $0.05$: the sPCE-optimised batch cannot commit
to an informative measurement direction without observed responses. DABS does
not lead at the smallest budgets. Multi-stage adaptive prunes candidates in
batches and scores higher while the posterior is still uninformative, an
advantage it does not sustain once the budget grows.

At $T = 8$, DABS reaches F1 $= 0.45$ and leads every baseline for the first time,
though its margin over multi-stage adaptive is within one standard error. From
$T = 10$ onward, DABS, the static design and the myopic baseline are
statistically indistinguishable: at $T = 16$, they reach $0.78$, $0.80$ and
$0.76$ respectively with overlapping CIs. The gap therefore closes before the
nominal boundary $T = p = 15$, consistent with the caveat in Section~3.6 that
sparsity can allow the active set to be resolved from fewer runs than the
nominal threshold. Multi-stage adaptive plateaus near $0.50$ from $T = 12$ and
Plackett-Burman reaches $0.49$ at its native $T = 16$, so both fall behind the
three methods that use the full posterior.

The comparison with Plackett--Burman and the multi-stage adaptive baseline measures the value of the Bayesian modelling framework, whereas the comparison with the myopic baseline separates that from the effect of amortised long-horizon planning.

This convergence matches the wedge prediction of Section~3.6. Timing and a single-rollout visualisation are given in Appendices~\ref{apE} and~\ref{apK}. The same pattern appears in the next two case studies (Sections~5.3 and~5.4).
\subsection{Scaling with the number of factors (HSC expansion benchmark)}
This second case study considers the ex vivo expansion of human hematopoietic stem cells (HSCs) from umbilical cord blood \citep{s3}, a clinically important process bottlenecked by cord-blood availability. The yield depends on the composition of the culture medium, which combines serum substitutes (bovine serum albumin, insulin, transferrin) with ten candidate cytokines including stem cell factor (SCF), Flt-3 ligand (FL), interleukin-3 (IL-3), and thrombopoietin (TPO). Cytokines are expensive to procure and biologically active, so both false positives and false negatives carry real cost. \citet{s3} used a $2^{10-6}$ fractional factorial design with $T = 16$ runs to screen all ten candidates and identified four cytokines (SCF, FL, IL-3, TPO) as the dominant drivers of expansion, later confirmed by follow-up optimisation studies. Full simulation details are in Appendix~\ref{apE}.

We simulate data from the spike-and-slab model of Section~3 with $\tau_{\text{active}} = 1.0$, $\tau_{\text{inactive}} = 0.01$,
$\alpha_0 = 3.0$, $\beta_0 = 1.0$, $\rho \sim \mathrm{Beta}(4,6)$ (prior mean $0.4$), and strong heredity on the
interaction indicators with $\pi_{\text{int}} = 0.5$.
At the fixed budget $T = 16$ of the original Yao et al.\ design, we vary $p \in \{10, 12, 14, 15, 16, 18, 20, 22, 24, 26\}$. The expected number of active main effects grows from 4 at $p = 10$ to 10.4 at $p = 26$, and the budget crosses the number of candidate factors at $p = T = 16$, the boundary between the identifiable and under-identified regimes. The sweep stops at $p = 26$ because the sufficient-statistics encoder, whose input dimension grows as $\mathcal{O}(p^4)$, exceeds our GPU memory budget beyond this point. Each policy is evaluated over 100 rollouts on main-effect F1 and FDR. TPR curves are reported in Appendix~\ref{app:case2_extra}.

The four baselines are as defined in Section~5.1. The 16-run Plackett-Burman (PB) design is only applicable for $p \leq 15$ since a $T$-run PB design requires $T \geq p + 1$. Consequently, no PB results are reported for $p \geq 16$. The static, DABS, myopic and multi-stage adaptive methods remain applicable across the full sweep.
\begin{figure}[!h]
\centering
\begin{tabular}{cc}
\subf{\includegraphics[width=78mm]{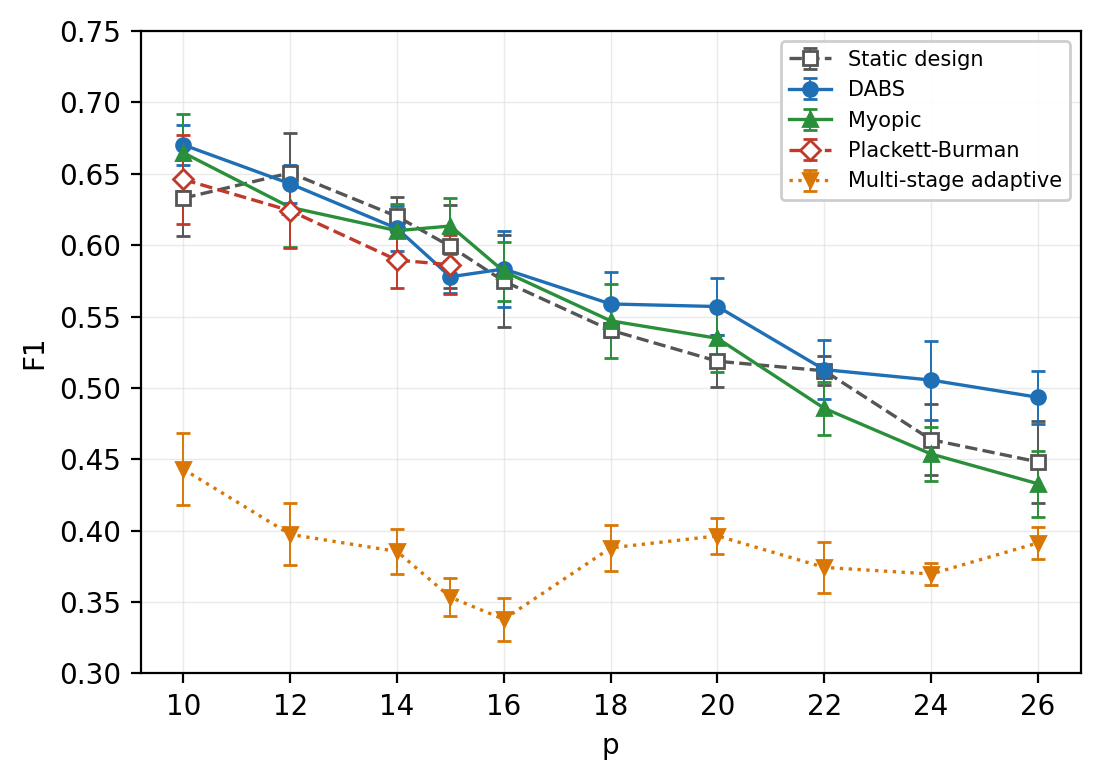}}
      {} &
\subf{\includegraphics[width=78mm]{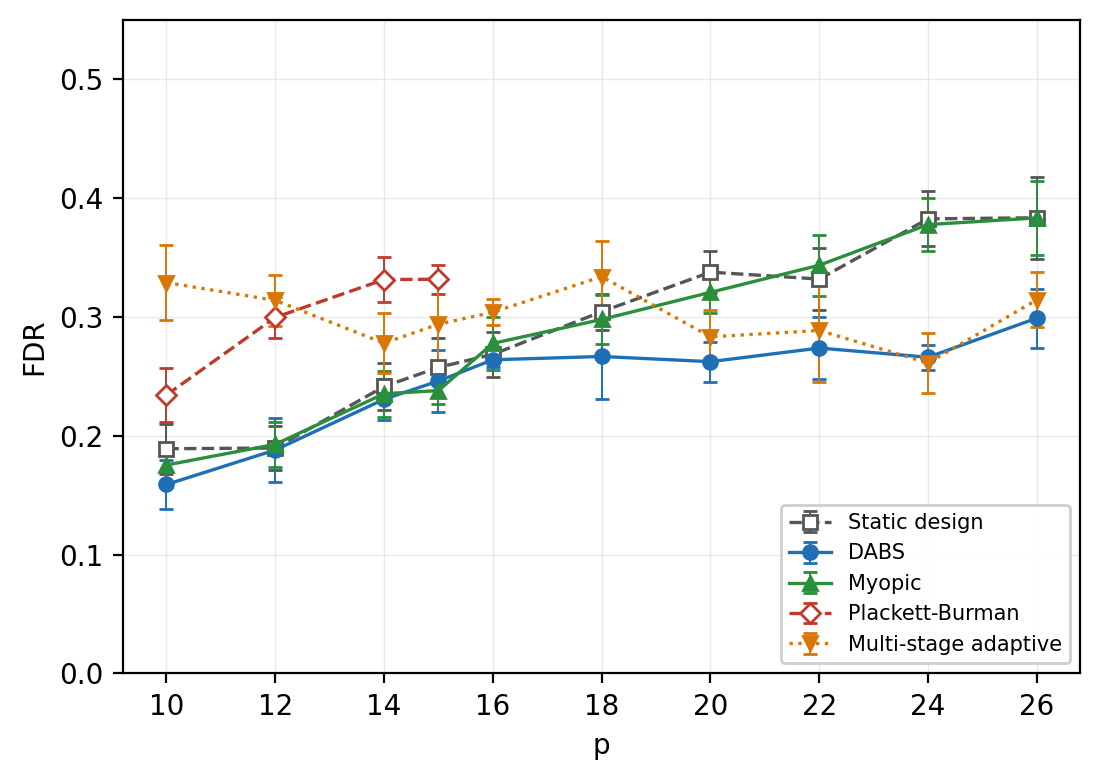}}
      {}
\end{tabular}
\caption{Main-effect F1 (left) and FDR (right) versus $p$ at $T = 16$. Bands are 95\% CIs over 5 seeds.}
\label{fig:c2_metrics}
\end{figure}
Figure~\ref{fig:c2_metrics} (left) shows main-effect F1 as a function of $p$. Below the boundary $p \leq 16$, DABS, the static design and the myopic baseline
are statistically indistinguishable, spanning $0.58$ to $0.67$, $0.57$ to $0.65$
and $0.58$ to $0.66$ respectively, matching the wedge prediction that no
adaptive gain remains when $T \geq p$. Plackett-Burman, applicable only up to
$p = 15$, spans $0.59$ to $0.65$ over its range.
Multi-stage
adaptive plateaus around $0.34$ to $0.44$. From $p = 18$ onwards the picture
changes: DABS sits above every baseline, reaching F1 $= 0.49$ at $p = 26$
against $0.45$ (static), $0.43$ (myopic), and $0.39$ (multi-stage adaptive).
Multi-stage adaptive stays flat
throughout, unable to convert additional budget into wider active-set search.
The pattern mirrors the wedge from Case Study 1: the adaptive gain of DABS opens
when the design is structurally information-starved, which here corresponds to
$p > T$.

Figure~\ref{fig:c2_metrics} (right) shows main-effect FDR. In the identifiable regime $p \leq 15$, DABS and the myopic baseline are the two
lowest curves, spanning $0.16$ to $0.25$ and $0.18$ to $0.24$ respectively,
ahead of the static design ($0.19$ to $0.26$), Plackett-Burman ($0.23$ to
$0.33$) and multi-stage adaptive ($0.28$ to $0.33$). 
From $p = 16$ onwards the methods separate. DABS' FDR flattens around $0.26$ to
$0.30$, while the static baseline climbs from $0.27$ to $0.38$ and the myopic
baseline from $0.28$ to $0.38$. At $p = 26$, DABS achieves an FDR of $0.30$
against $0.38$ for both, an eight-percentage-point improvement. Multi-stage
adaptive stays flat between $0.26$ and $0.33$ across the sweep.

The mechanism is asymmetric. The static design must distribute its budget uniformly across all $p$ candidates because it has no information about which ones are active. As $p$ grows, each coordinate receives less measurement effort and the posterior becomes weakly informative for a larger fraction of candidates. The Gibbs decoder, applied with threshold $0.5$, then commits to false positives among the uncertain inactives. DABS instead uses early observations to suppress probing on coordinates that show no evidence of activity, and concentrates the remaining budget on the suspected active subset. The FDR gap over the static baseline drives the F1 gap of Figure~\ref{fig:c2_metrics} (left).

Together these two case studies characterise the screening-regime advantage of DABS along complementary axes, sweeping $T$ at fixed $p$ and sweeping $p$ at fixed $T$. Both confirm that the advantage of DABS over the matched Bayesian controls opens when the design is structurally information-starved.
\subsection{Two-factor interaction screening (metallurgical benchmark)}
This third case study considers the electrochemical dissolution of GH4738 nickel-based superalloy scrap \citep{s6}. The process is governed by $p = 8$ parameters, of which four are the active main effects: current density ($X_1$), H$_2$SO$_4$ concentration ($X_2$), NiCl$_2$ concentration ($X_4$) and electrolysis time ($X_6$). Prior analysis \citep{s6} identifies three significant two-factor interactions, $X_1 X_6$, $X_2 X_4$, and $X_2 X_6$, all satisfying strong heredity. The original study used a 12-run Plackett-Burman design. Even at $T = 16$, the interaction-recovery problem is under-identified: a Resolution V fractional factorial requires $2^{8-2} = 64$ runs to estimate the $p + \binom{p}{2} = 36$ terms without aliasing and a modern D-optimal design requires $37$ runs \citep{b3}. The identifiability boundary of Section~3.6 is therefore $T = 36$ at $p = 8$, and every budget tested lies below it.

We simulate data from the spike-and-slab model of Section~3 with $\tau_{\text{active}} = 2.0$, $\tau_{\text{inactive}} = 0.01$, $\alpha_0 = 3.0$,
$\beta_0 = 0.25$, and $\rho$ fixed at $0.5$.
Strong heredity is enforced on the interaction indicators, with
$\pi_{\mathrm{int}} = 0.5$ throughout. We sweep the budget $T$ at three problem sizes: $T \in \{3, 4, 5, 6, 8, 10, 12, 16, 18, 20\}$ at $p = 4$ and $T \in \{6, 8, 10, 12, 14, 16, 18, 20, 24\}$ at $p \in \{6, 8\}$. 
The largest budgets remain well below the identifiability boundary of $T = 36$. Each policy is evaluated over 100 rollouts on main-effect F1 and interaction F1. TPR and FDR curves for both main effects and interactions are reported in Appendix~\ref{app:case3_extra}.

The four baselines are as defined in Section~5.1. Two are omitted from this case study. Plackett-Burman is a main-effects-only design whose interaction columns are fully aliased and cannot resolve two-factor interactions. The multi-stage adaptive baseline is likewise main-effects-only. This case study therefore compares DABS against the two baselines whose designs can resolve two-factor interactions: the static sPCE-optimised design and the myopic baseline.
\begin{figure}[!h]
\centering
\includegraphics[width=\linewidth]{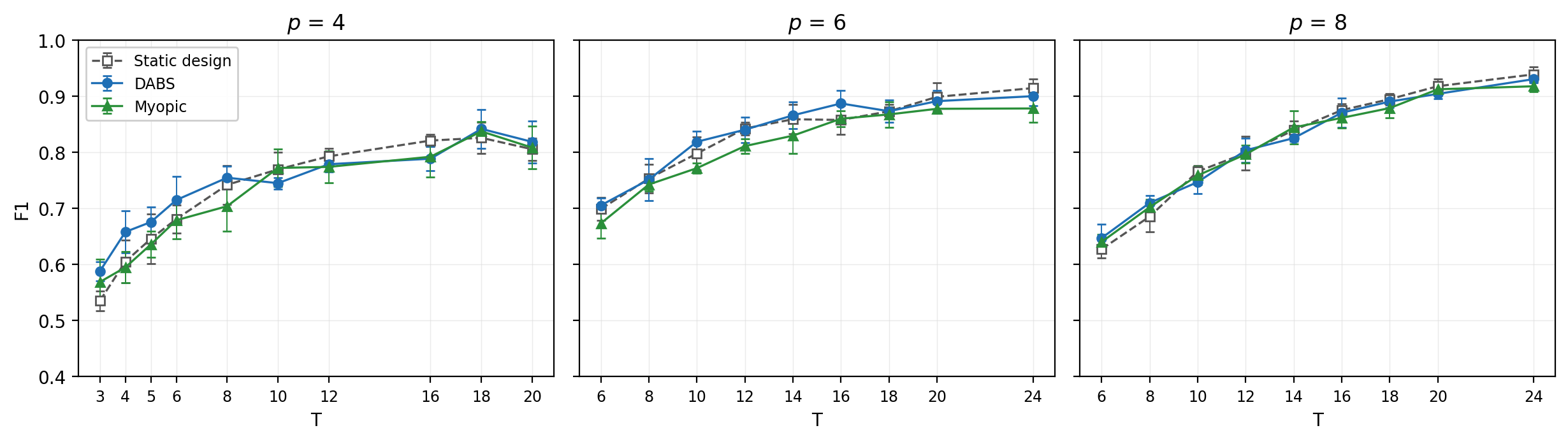}
\caption{Main-effect F1 versus budget $T$ across problem sizes $p \in \{4, 6, 8\}$. Bands are 95\% CIs over 5 seeds.}
\label{fig:c3_F1_main}
\end{figure}
Figure~\ref{fig:c3_F1_main} shows main-effect F1 across the budget sweep. At
every problem size $p \in \{4, 6, 8\}$ the three policies track each other
closely. DABS and the static design differ by at most 3 percentage points at
$p = 6$ and $p = 8$, and the myopic baseline stays within 6 percentage points of
the better of the two. All three climb monotonically with $T$ toward a common
ceiling (F1 $\approx 0.83$ at $p = 4$, $0.91$ at $p = 6$, $0.94$ at $p = 8$).
The one clear separation sits at the smallest budgets, where the design is
under-identified for main effects. At $p = 4$, $T = 3$, DABS reaches $0.59$
against $0.54$ for the static design, with non-overlapping confidence intervals,
consistent with the wedge prediction from the previous case studies. Main-effect
recovery at these budgets is otherwise a ceiling story, and the policies
separate only where the budget is too small to identify the main effects at all.
\begin{figure}[!h]
\centering
\includegraphics[width=\linewidth]{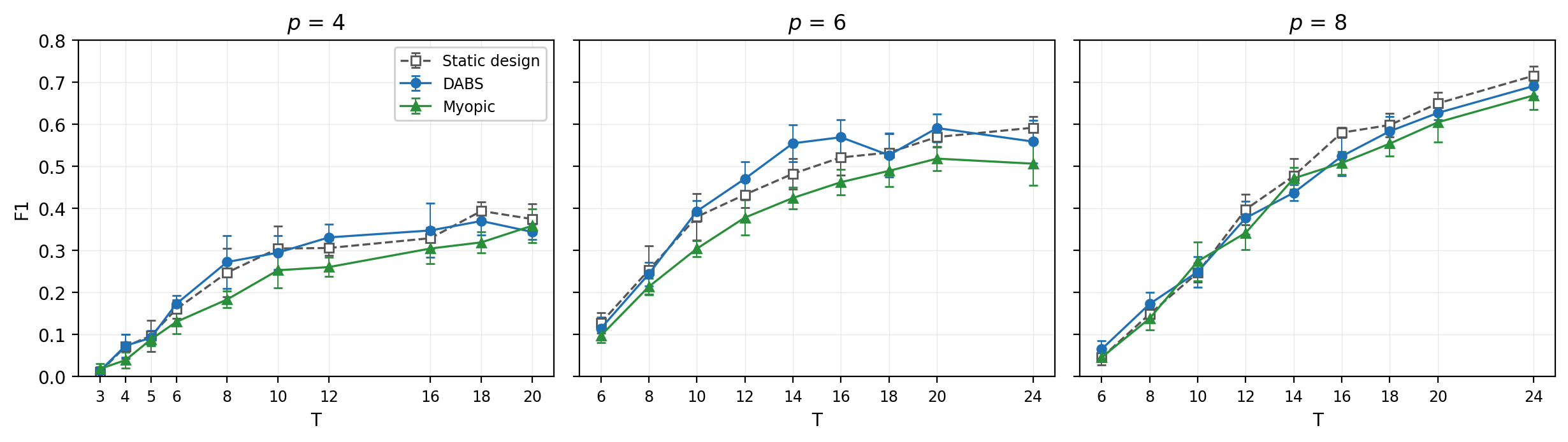}
\caption{Two-factor interaction F1 versus budget $T$ across problem sizes $p \in \{4, 6, 8\}$. Bands are 95\% CIs over 5 seeds.}
\label{fig:c3_F1_int}
\end{figure}
Figure~\ref{fig:c3_F1_int} shows interaction F1. All three policies climb from
near-zero at the smallest budgets toward a $p$-dependent plateau at high $T$. At
the top of each sweep DABS and the static design reach $0.34$ and $0.37$ at
$p = 4$, $0.56$ and $0.59$ at $p = 6$, and $0.69$ and $0.72$ at $p = 8$. The
myopic baseline reaches $0.36$, $0.51$ and $0.67$. The three plateaus are within
noise of each other at $p = 4$ and $p = 8$, so the budgets tested do not
separate the policies on interaction recovery at the top of the sweep.

The ordering DABS $>$ static $>$ myopic appears at $p = 6$ in the mid-budget
window. At $T = 14$, DABS reaches interaction F1 $= 0.55$ against $0.48$ (static)
and $0.43$ (myopic), with non-overlapping confidence intervals between DABS and
the static design. The advantage persists at $T = 12$ and $T = 16$ with
overlapping intervals and closes by $T = 24$. It does not appear at the other
two problem sizes. At $p = 4$ and $p = 8$ DABS and the static design are
statistically indistinguishable across the sweep, and there is no problem size
and budget in this case study at which the static design leads significantly. The mechanism is a two-phase strategy, in
which DABS uses early experiments to identify candidate active main effects and
later experiments to probe the heredity-consistent $\xi_i \xi_j$ pairs among
them. Under strong heredity the interaction search space can be pruned only once
the active parents are known, so the gain sits in the budget window between main-effect
resolution and interaction saturation. The grids of this case study span that
window only at $p = 6$. At $p = 4$ the window is a few runs wide, and at
$p = 8$ the largest budget tested, $T = 24$, still lies well below the
identifiability boundary of $T = 36$, so the sweep never reaches the saturated
end.

These results extend the wedge prediction beyond main-effect screening. The gain
appears where the mechanism predicts it and is absent only where the budget grid
does not reach that window. Across all three problem sizes DABS matches the
static design on interaction recovery, a design that is re-optimised for every
$(p, T)$. DABS is also trained offline once and generates each design in a single forward pass, whereas sequential adaptive procedures such as the myopic baseline solve a fresh optimisation at every step, which limits their use in time-sensitive workflows.
\subsection{Ablations, Robustness \& Calibration}
We assess the main design choices in DABS and its robustness to misspecification. We ablate the logit-$\ell_2$ penalty that prevents policy collapse, the sufficient-statistics encoder against a set-equivariant encoder, the policy initialisation, the reward structure, and the training objective $I(z; h_T)$ against the total-information target $I(\theta; h_T)$ used by DAD-style methods. We further evaluate robustness to the observation noise level, to a misspecified sparsity prior, and to a heredity mismatch in which a policy trained under strong heredity is evaluated under weak heredity. A broader hyperparameter sensitivity analysis completes the study. Results and discussion are provided in Appendix~\ref{apF}. 

We also assess whether the credible intervals produced at deployment are
calibrated. They inherit the quality of the screening decision. At the $95\%$
level, main-effect intervals are close to nominal for factors whose activity is
correctly identified, reaching $0.926$, $0.916$ and $0.934$ across the three
case studies, and are conservative for coefficients correctly ruled out.
Coverage on active two-factor interactions is lower, $0.753$ in Case Study~3,
so interaction intervals at tight budgets should be treated as indicative
rather than calibrated. Appendix~\ref{apI} reports the full calibration
study.
\subsection{Limitations}
First, training DABS requires substantial upfront computation, particularly for large $p$ and $T$, where the Student-$t$ covariance scales as $T^2$ per contrastive sample. This cost is amortised, since a trained policy is reused across new instances under the same prior, and it is controlled through the number of training steps and contrastive samples. Second, the closed-form Student-$t$ marginal relies on conjugacy between the spike-and-slab prior and the Gaussian likelihood. Extending DABS to non-conjugate priors or non-Gaussian responses would require likelihood-free estimators or auxiliary-variable augmentations. Finally, the method assumes strong heredity.
Its scaling to large $p$ is limited by the sufficient-statistics encoder used in
the large-$p$ sweeps, whose input dimension grows as $\mathcal{O}(p^4)$. This
sets the range demonstrated here, up to $p = 26$ for main-effect screening and
$p = 8$ when two-factor interactions are included. Appendix~\ref{apF} shows that
the alternative encoder avoids this growth, which makes it the more promising
route to larger $p$.
\section{Conclusions}
In this paper we introduced DABS, a policy-based adaptive Bayesian experimental design method for discrete factorial screening under spike-and-slab priors. By exploiting conjugacy between the prior and likelihood, DABS trains a single policy via Prior Contrastive Estimation on an analytical Student-$t$ marginal and recovers the active factor set through posterior inference at deployment. In our experiments, DABS matches or outperforms the strongest non-adaptive Bayesian baseline on a biotechnological screening benchmark, a scaling study with varying dimensionality, and an interaction-recovery benchmark under strong heredity. The largest gains arise precisely where classical orthogonal
designs fall short. In the scaling study DABS holds the lowest false discovery rate at
eight of the ten problem sizes tested, and its margin over the static design
widens as $p$ grows. At $p = 26$ DABS reaches a false discovery rate of $0.30$
against $0.38$ for the static design. 
DABS thereby bridges the gap between modern amortised BOED methods and classical screening practice, enabling efficient adaptive design that directly targets mutual information through a closed-form marginal over the support variables.
\newpage
\bibliography{main}

@InProceedings{ref1,
  title = 	 {\text{Deep Adaptive Design}: Amortizing \text{Sequential Bayesian Experimental Design}},
  author =       {Foster, Adam and Ivanova, Desi R and Malik, Ilyas and Rainforth, Tom},
  booktitle = 	 {International Conference on Machine Learning},
  pages = 	 {3384--3395},
  year = 	 {2021}
}

@inproceedings{ref2,
  title={\text{Categorical Reparameterization with Gumbel-Softmax}},
  author={Jang, Eric and Gu, Shixiang and Poole, Ben},
  booktitle={International Conference on Learning Representations},
  year={2017}
}

@article{ref4,
author = {Ishwaran, Hemant and Rao, J.},
year = {2005},
pages = {730-773},
title = {\text{Spike and Slab Variable Selection}: Frequentist and \text{Bayesian} strategies},
volume = {33},
journal = {The Annals of Statistics}
}

@article{ref5,
  author  = {Chaloner, Kathryn and Verdinelli, Isabella},
  title   = {\text{Bayesian Experimental Design: A Review}},
  journal = {Statistical Science},
  volume  = {10},
  number  = {3},
  pages   = {273--304},
  year    = {1995}
}

@InProceedings{ref6,
  title = 	 {\text{Implicit Deep Adaptive Design}: \text{Policy-Based Experimental Design} without Likelihoods},
  author =       {Ivanova, Desi R and Foster, Adam and Kleinegesse, Steven and Gutmann, Michael U. and Rainforth, Tom},
  booktitle = 	 {Advances in Neural Information Processing Systems},
  pages = 	 {25785--25798},
  year = 	 {2021}
}

@article{ref7,
  author  = {Box, George E.P. and Meyer, R. Daniel},
  title   = {\text{An analysis for Unreplicated Fractional Factorials}},
  journal = {Technometrics},
  volume  = {28},
  number  = {1},
  pages   = {11--18},
  year    = {1986}
}

@article{ref8,
  author  = {George, Edward I. and McCulloch, Robert E.},
  title   = {\text{Variable Selection Via Gibbs Sampling}},
  journal = {Journal of the American Statistical Association},
  volume  = {88},
  number  = {423},
  pages   = {881--889},
  year    = {1993}
}

@InProceedings{ref10,
  author  = {Maddison, Chris J. and Mnih, Andriy and Teh, Yee Whye},
  title   = {The Concrete Distribution: A Continuous Relaxation of Discrete Random Variables},
  booktitle = {International Conference on Learning Representations},
  year    = {2017}
}

@article{ref12,
  author  = {Geman, Stuart and Geman, Donald},
  title   = {Stochastic Relaxation, \text{Gibbs Distributions}, and the \text{Bayesian Restoration} of Images},
  journal = {IEEE Transactions on Pattern Analysis and Machine Intelligence},
   volume = {PAMI-6},
   number = {6},
  year    = {1984}
}

@article{ref14,
  author  = {Hamada, M. and Wu, C. F. Jeff},
  title   = {Analysis of designed experiments with complex aliasing},
  journal = {Journal of Quality Technology},
   volume = {24},
   number = {3},
   pages  = {130--137},
  year    = {1992}
}

@book{ref15,
    author = {Montgomery, Douglas C.},
    title = {Design and Analysis of Experiments},
    publisher = {Wiley},
    year = {2017}
}

@book{ref17,
    author = {Box, George E.P. and Hunter, J. Stuart and Hunter, William G.},
    title = {Statistics for experimenters: Design, Innovation, and Discovery},
    publisher = {Wiley},
    year = {2005}
}

@article{ref18,
  author  = {Ryan, Elizabeth G. and Drovandi, Christopher C. and McGree, James M. and Pettitt, Anthony N.},
  title   = {A review of \text{Modern Computational Algorithms for Bayesian Optimal Design}},
  journal = {International Statistical Review},
   volume = {84},
   number = {1},
    pages   = {128--154},
  year    = {2016}
}

@article{ref19,
  author  = {Lindley, D.V.},
  title   = {On a measure of the information provided by an experiment},
  journal = {The Annals of Mathematical Statistics},
     volume = {27},
   number = {4},
    pages   = {986--1005},
  year    = {1956}
}

@InProceedings{ref20,
  title = 	 {\text{On Nesting Monte Carlo Estimators}},
  author  = {Rainforth, Tom and Cornish, Robert and Yang, Hongseok and Warrington, Andrew and Wood, Frank},
  booktitle = {International Conference on Machine Learning},
  year      = {2018},
  pages     = {4267--4276}
}

@inproceedings{ref21,
  title     = {A \text{Unified Stochastic Gradient Approach to Designing Bayesian-Optimal Experiments}},
  author    = {Foster, Adam and Jankowiak, Martin and O'Meara, Matthew J. and Teh, Yee Whye and Rainforth, Tom},
  booktitle = {International Conference on Artificial Intelligence and Statistics},
  year      = {2019},
  pages     = {2959--2968}
}

@article{ref23,
  title   = {An Algorithm for the Construction of {D-Optimal} Experimental Designs},
  author  = {Mitchell, Toby J.},
  journal = {Technometrics},
  volume  = {16},
  number  = {2},
  pages   = {203--210},
  year    = {1974}
}

@book{ref24,
    author = {Wu, C.F. Jeff and Hamada, Michael S.},
    title = {Experiments: Planning, Analysis, and Optimization},
    publisher = {Wiley},
    year = {2009}
}

@article{ref25,
  title   = {Bayesian Variable Selection with Related Predictors},
  author  = {Chipman, Hugh},
  journal = {The Canadian Journal of Statistics},
  volume  = {24},
  number  = {1},
  pages   = {17--36},
  year    = {1996}
}

@inproceedings{ref27,
  title     = {\text{REBAR}: low-variance, unbiased gradient estimates for discrete latent variable models},
  author    = {Tucker, George and Mnih, Andriy and Maddison, Chris J. and Lawson, Dietrich and Sohl-Dickstein, Jascha},
  booktitle = {Advances in Neural Information Processing Systems},
  year      = {2017},
  pages     = {2624--2633}
}

@article{ref30,
  title   = {Approaches for \text{Bayesian Variable Selection}},
  author  = {George, Edward I. and McCulloch, Robert E.},
  journal = {Statistica Sinica},
  volume  = {7},
  pages   = {339--373},
  year    = {1997}
}

@article{ref31,
  title   = {The Practical Implementation of \text{Bayesian Model Selection}},
  author  = {Chipman, Hugh and George, Edward I. and McCulloch, Robert E.},
  journal = {Lecture Notes-Monograph Series},
  volume  = {38},
  pages   = {65--116},
  year    = {2001},
  publisher = {Institute of Mathematical Statistics}
}

@article{RW2,
author = {Jacob Abugre Akubire and Byran J. Smucker},
title = {A comparison of supersaturated designs and orthogonal arrays},
journal = {Journal of Statistical Computation and Simulation},
volume = {95},
number = {9},
pages = {1928--1943},
year = {2025},
publisher = {Taylor \& Francis},
}

@article{RW3,
  title={Casting a Spotlight on Factorial Design: Exploring the Power of \text{DoE} for Experiment Screening and Optimization: A Review},
  author={Prathap Madeswara Guptha and Vijayaraj Surendran and Raghavendra Kumar Gunda},
journal = {Indian Journal of Pharmaceutical Education and Research},
volume = {59},
number = {2},
pages = {406--416},
year = {2025},
publisher = {Taylor \& Francis},
}

@article{RW6,
  title={Bayesian Variable Selection in Linear Regression},
  author={Mitchell, T.J. and Beauchamp, J.J.},
journal = {Journal of the American Statistical Association},
volume = {83},
number = {404},
pages = {1023--1032},
year = {1988}
}

@article{RW7,
  title={A review of Bayesian variable selection methods: what, how and which},
  author={O'Hara, R.B. and Sillanpää, M.J.},
journal = {\text{Bayesian Analysis}},
volume = {4},
pages = {85--117},
year = {2009}
}

@article{RW8,
  title={Bayesian iterative screening in ultra-high dimensional linear regressions},
  author={Wang, Run and Dutta, Somak and Roy, Vivekananda},
journal = {Bayesian Analysis},
note = {Advance Publication},
pages = {1--26},
year = {2025}
}

@article{RW9,
  title={Double-robust \text{Bayesian} variable selection and model prediction with spherically symmetric error},
  author  = {Ouyang, Linhan and Yan, Ling and Liu, Lijun and Sun, Minghe and Wang, Min},
journal = {IISE Transactions},
volume = {58},
number = {2},
pages = {181--194},
year = {2026}
}

@InProceedings{RW11,
  title = 	 {Step-{DAD}: Semi-Amortized Policy-Based {B}ayesian Experimental Design},
  author =       {Hedman, Marcel and Ivanova, Desi R. and Guan, Cong and Rainforth, Tom},
  booktitle =  {International Conference on Machine Learning},
  pages = 	 {22904--22923},
  year = 	 {2025},
  volume = 	 {267}
}

@InProceedings{RW13,
  title = 	 {Mutual information estimation via normalizing flows},
  author =       {Butakov, Ivan and Tolmachev, Alexander and Malanchuk, Sofia and Neopryatnaya, Anna and Frolov, Alexey},
  booktitle = {Advances in Neural Information Processing Systems},
  year      = {2024},
  volume = {37},
  pages = {3027--3057}
}

@ARTICLE{RW17,
  title  = "\text{Pre-trained Gaussian processes for Bayesian optimization}",
  author = "Wang, Zi and Dahl, George E and Swersky, Kevin and Lee, Chansoo and Nado, Zachary and Gilmer, Justin and Snoek, Jasper and Ghahramani, Zoubin",
  journal  = "Journal of Machine Learning Research",
  volume   =  25,
  pages    = "1--83",
  year     =  2024,
}

@article{RW18,
author = {Guo, Dan and Wang, Xilu and Gao, Kailai and Jin, Yaochu and Ding, J.L. and Chai, Tianyou},
year = {2021},
pages = {1-14},
title = {\text{Evolutionary} \text{Optimization of High-Dimensional Multiobjective} and \text{Many-Objective Expensive Problems} Assisted by a \text{Dropout Neural Network}},
volume = {PP},
journal = {IEEE Transactions on Systems, Man, and Cybernetics: Systems},
}

@article{RW21,
  author        = {Huang, Daolang and Guo, Yujia and Acerbi, Luigi and Kaski, Samuel},
  title         = {\text{Amortized Bayesian} \text{Experimental Design for Decision-Making}},
  year          = {2024},
  eprint        = {2411.02064v2},
  archivePrefix = {arXiv},
  primaryClass  = {stat.ML},
  journal       = {arXiv preprint arXiv:2411.02064}
}

@article{RW22,
  author        = {Strouwen, Arno and Micluta-C{\^a}mpeanu, Sebastian},
  title         = {\text{Deep Adaptive Model-Based Design} of \text{Experiments}},
  year          = {2026},
  eprint        = {2603.16146v2},
  archivePrefix = {arXiv},
  primaryClass  = {stat.ML},
  journal       = {arXiv preprint arXiv:2603.16146}
}

@article{RW23,
  author        = {Fanuel, Michaël and Schreurs, Joachim and Suykens, Johan A.K.},
title = {Diversity Sampling is an Implicit Regularization for Kernel Methods},
journal = {SIAM Journal on Mathematics of Data Science},
volume = {3},
number = {1},
pages = {280-297},
year = {2021}
}

@InProceedings{RW24,
author={Schreurs, Joachim
and De Meulemeester, Hannes
and Fanuel, Micha{\"e}l
and De Moor, Bart
and Suykens, Johan A. K.},
title={\text{Leverage Score Sampling for Complete Mode Coverage in Generative Adversarial} \text{Networks}},
booktitle={Machine Learning, Optimization, and Data Science},
year={2022},
pages={466--480}
}

@article{s2,
title = {A study of inulinase production in \text{Aspergillus} niger using fractional factorial design},
journal = {Bioresource Technology},
volume = {54},
number = {3},
pages = {315-320},
year = {1995},
author = {Poorna, V. and Kulkarni, P.R.}
}

@article{s3,
title = {A systematic strategy to optimize ex vivo expansion medium for human hematopoietic stem cells derived from umbilical cord blood mononuclear cells},
journal = {Experimental Hematology},
volume = {32},
number = {8},
pages = {720-727},
year = {2004},
author = {Chao-Ling Yao and I-Ming Chu and Tzu-Bou Hsieh and Shiaw-Min Hwang},
}

@Article{s6,
AUTHOR = {Liu, Guiqun and Fang, Xinyu and Zhang, Xiaoli and Lv, Guanglei},
TITLE = {Optimization of Dissolution Parameters for GH4738 Scrap via Response Surface Methodology},
JOURNAL = {Materials},
VOLUME = {18},
YEAR = {2025},
NUMBER = {4},
ARTICLE-NUMBER = {793},
}

@InProceedings{RW30,
  title = 	 {\text{Prediction-Oriented Bayesian Active Learning}},
  author =       {Bickford Smith, Freddie and Kirsch, Andreas and Farquhar, Sebastian and Gal, Yarin and Foster, Adam and Rainforth, Tom},
  booktitle = 	 {International Conference on Artificial Intelligence and Statistics},
  pages = 	 {7331--7348},
  year = 	 {2023},
  volume = 	 {206},
  month = 	 {25--27 Apr},
}

@InProceedings{RW31,
  title = 	 {{CO}-{BED}: Information-Theoretic Contextual Optimization via {B}ayesian Experimental Design},
  author =       {Ivanova, Desi R. and Jennings, Joel and Rainforth, Tom and Zhang, Cheng and Foster, Adam},
  booktitle = 	 {International Conference on Machine Learning},
  pages = 	 {14445--14464},
  year = 	 {2023},
  volume = 	 {202},
  month = 	 {23--29 Jul},
}

@article{RW32,
  title     = {Efficient global optimization of expensive black-box functions},
  author    = {Jones, Donald R. and Schonlau, Matthias and Welch, William J.},
  journal   = {Journal of Global Optimization},
  volume    = {13},
  number    = {4},
  pages     = {455--492},
  year      = {1998},
}

@InProceedings{RW33,
  title  = "\text{Variational Bayesian Optimal Experimental Design}",
  author = "Foster, Adam and Jankowiak, Martin and Bingham, Eli and Horsfall,
            Paul and Teh, Yee Whye and Rainforth, Tom and Goodman, Noah",
  year   =  2019,
  booktitle = 	 {Advances in Neural Information Processing Systems},
  volume = 	 {32}
}

@book{b1,
    author = {Pukelsheim, Friedrich},
    title = {Optimal Design of Experiments},
    publisher = {Wiley},
    year = 1993
}

@ARTICLE{b2,
  title   = "General equivalence theory for optimum designs (approximate
             theory)",
  author  = "Kiefer, J.",
  journal = "Annals of Statistics",
  volume  =  2,
  number  =  5,
  pages   = "849--879",
  year    =  1974
}

@book{b3,
    author = {Goos, Peter and Jones, Bradley},
    title = {Optimum Design of Experiments: A Case Study Approach},
    publisher = {Wiley},
    year = {2011}
}

@ARTICLE{b4,
  title   = {Efficient space-filling and non-collapsing sequential design strategies for simulation-based modeling},
  author  = {Crombecq, K. and Laermans, E. and Dhaene, T.},
  journal = "European Journal of Operational Research",
  volume  =  214,
  pages   = "683--696",
  year    =  2011
}

@InProceedings{b5,
  title  = "\text{Experimental Design on a Budget for Sparse Linear Models and Applications}",
  author = 	 {Ravi, Sathya Narayanan and Ithapu, Vamsi and Johnson, Sterling and Singh, Vikas},
  booktitle = 	 {International Conference on Machine Learning},
  pages = 	 {583--592},
  year = 	 {2016},
  volume = 	 {48}
}

@ARTICLE{pw2,
  title   = {Distilled Sensing: Adaptive Sampling for Sparse Detection and Estimation},
  author  = {Haupt, Jarvis and Castro, Rui M. and Nowak, Robert},
  journal = "IEEE Transactions on Information Theory",
  volume = 57,
  number  =  9,
  year    =  2011
}

@ARTICLE{ref16,
  title   = {The Design of Optimum Multifactorial Experiments},
  author  = {Plackett, R.L. and Burman, J.P.},
  journal = "Biometrika",
    volume = 33,
  number  =  4,
  pages   = "305--325",
  year    =  1946
}
\newpage
\appendix
\section{Proof of Proposition 1} \label{apAa}
Proposition 1 follows from classical convexity arguments in optimal design~\citep{b1, b2}, but we include a proof for completeness and to explicitly adapt the argument to our setting.

The design space $\mathcal{X} = [-1, 1]^p$ is a convex polytope whose extreme points are the vertices $\{-1, +1\}^p$. In the linear--Gaussian setting, the mutual information objective can be written as
\begin{equation}
f(\xi) = \tfrac{1}{2}\log\left(1 + \frac{\xi^\top \Sigma \xi}{\sigma^2}\right),
\end{equation}
where $\Sigma \succeq 0$, which equals the prior covariance $\Sigma_\beta$ for a single
observation and more generally the current posterior covariance of $\beta$. Since
$t \mapsto \tfrac{1}{2}\log(1 + t/\sigma^2)$ is strictly increasing maximising $f$ over $\mathcal{X}$ is equivalent to maximising the quadratic form $q(\xi) := \xi^\top \Sigma \xi$.

We now show that $q(\xi)$ is maximised over $\mathcal{X}$ at a vertex. Fix all coordinates of $\xi$ except $\xi_i$. The restriction of $q$ to $\xi_i$ is
\begin{equation}
q(\xi_i) = a\, \xi_i^2 + b\, \xi_i + c,
\end{equation}
where $a = \Sigma_{ii} \geq 0$, $b = 2 \sum_{j \neq i} \Sigma_{ij}\, \xi_j$, and $c$ is independent of $\xi_i$. This is a convex quadratic function of $\xi_i$ over the interval $[-1, 1]$, so its maximum is attained at the boundary $\xi_i \in \{-1, +1\}$.

Applying this argument sequentially to each coordinate shows that the maximum of
$\xi^\top \Sigma \xi$ over $[-1, 1]^p$ is attained at some point of $\{-1, +1\}^p$. Since $f(\xi)$ is a monotone increasing function of $\xi^\top \Sigma \xi$, the same holds for $f(\xi)$.
\newpage
\section{Ablations and robustness} \label{apF}
\subsection{Experimental Setup}
\label{sec:ablation-setup}
All ablations in this appendix are anchored on a shared problem class so that effects are directly comparable. We consider $p = 10$ candidate factors, budget $T = 8$, observation noise $\sigma = 0.5$ and sparsity prior $\rho \sim \mathrm{Beta}(2, 8)$ (prior mean 0.2), with $\tau_{\text{active}} = 2.0$, $\tau_{\text{inactive}} = 0.01$, $\alpha_0 = 3.0$ and $\beta_0 = 1.0$. Strong heredity is enforced on the interaction indicators, with $\pi_{\mathrm{int}} = 0.5$. Each policy uses the sufficient-statistics encoder of Section~3.4 and warm-starts from a static sPCE-optimised checkpoint before adaptive fine-tuning. Training runs for 30,000 steps with the recipe of Section~5.2. Each ablation cell is trained with 5 seeds, and the baseline configuration with 12. At evaluation, each policy is rolled out 50 times, decoded by the sampler of Appendix~\ref{apC} with 200 draws after 100 burn-in steps, with posterior inclusion threshold 0.5. We report main-effect F1 as the primary metric. Interaction F1 is near zero at this problem class because strong heredity combined with the Beta(2, 8) prior yields very few active interactions per instance. The interaction-recovery story is treated in Section~5.4 at a problem class where interactions are prevalent.
\subsection{Hyperparameter Sensitivity Analysis}
This subsection sweeps every real-valued hyperparameter of the training recipe one at a time around the baseline configuration and reports the resulting main-effect F1. Table~\ref{tab:hp_sensitivity} reports the 25 off-baseline cells. They fall into three groups. The optimisation group covers the learning rate, total number of training steps, exponential learning-rate decay and Adam weight decay. The capacity group covers the encoder hidden dimension and the encoding output dimension. The regularisation and Gumbel-Softmax group covers the logit-$\ell_2$ penalty, the diversity bonus, and the three Gumbel-Softmax annealing parameters $\tau_{\mathrm{start}}$, $\tau_{\mathrm{end}}$, and the anneal fraction. We report main-effect F1 with the standard error over seeds.
\begin{table}[h!]
\centering
\small
\renewcommand{\arraystretch}{1.1}
\begin{tabular}{@{}llc@{}}
\toprule
\textbf{Hyperparameter} & \textbf{Value} & \textbf{Main F1} \\
\midrule

\multirow{5}{*}{Learning rate}
 & $1 \times 10^{-4}$              & $0.300 \pm 0.032$ \\
 & $5 \times 10^{-4}$              & $0.312 \pm 0.032$ \\
 & $\mathbf{1 \times 10^{-3}}$     & $\mathbf{0.276 \pm 0.013}$ \\
 & $2 \times 10^{-3}$              & $0.308 \pm 0.028$ \\
 & $5 \times 10^{-3}$              & $0.277 \pm 0.028$ \\
\midrule

\multirow{3}{*}{Training steps}
 & 15{,}000                        & $0.270 \pm 0.020$ \\
 & \textbf{30{,}000}               & $\mathbf{0.276 \pm 0.013}$ \\
 & 60{,}000                        & $0.311 \pm 0.021$ \\
\midrule

\multirow{3}{*}{LR decay $\gamma$}
 & $0.90$                          & $0.312 \pm 0.030$ \\
 & $\mathbf{0.95}$                 & $\mathbf{0.276 \pm 0.013}$ \\
 & $0.98$                          & $0.308 \pm 0.029$ \\
\midrule

\multirow{3}{*}{Adam weight decay}
 & $\mathbf{0}$                    & $\mathbf{0.276 \pm 0.013}$ \\
 & $1 \times 10^{-4}$              & $0.260 \pm 0.016$ \\
 & $1 \times 10^{-3}$              & $0.305 \pm 0.007$ \\
\midrule

\multirow{3}{*}{Hidden dim}
 & $32$                            & $0.299 \pm 0.023$ \\
 & $\mathbf{64}$                   & $\mathbf{0.276 \pm 0.013}$ \\
 & $128$                           & $0.300 \pm 0.012$ \\
\midrule

\multirow{3}{*}{Encoding dim}
 & $16$                            & $0.298 \pm 0.022$ \\
 & $\mathbf{32}$                   & $\mathbf{0.276 \pm 0.013}$ \\
 & $64$                            & $0.297 \pm 0.029$ \\
\midrule

\multirow{4}{*}{Logit-$\ell_2$ weight $\lambda$}
 & $0.001$                         & $0.289 \pm 0.014$ \\
 & $\mathbf{0.01}$                 & $\mathbf{0.276 \pm 0.013}$ \\
 & $0.05$                          & $0.268 \pm 0.014$ \\
 & $0.1$                           & $0.311 \pm 0.009$ \\
\midrule

\multirow{3}{*}{Diversity penalty}
 & $\mathbf{0}$                    & $\mathbf{0.276 \pm 0.013}$ \\
 & $0.01$                          & $0.272 \pm 0.007$ \\
 & $0.1$                           & $0.283 \pm 0.018$ \\
\midrule

\multirow{3}{*}{$\tau_{\mathrm{start}}$}
 & $0.5$                           & $0.314 \pm 0.014$ \\
 & $\mathbf{1.0}$                  & $\mathbf{0.276 \pm 0.013}$ \\
 & $2.0$                           & $0.315 \pm 0.030$ \\
\midrule

\multirow{3}{*}{$\tau_{\mathrm{end}}$}
 & $0.1$                           & $0.289 \pm 0.042$ \\
 & $\mathbf{0.3}$                  & $\mathbf{0.276 \pm 0.013}$ \\
 & $0.5$                           & $0.286 \pm 0.038$ \\
\midrule

\multirow{3}{*}{Anneal fraction}
 & $0.3$                           & $0.294 \pm 0.011$ \\
 & $\mathbf{0.5}$                  & $\mathbf{0.276 \pm 0.013}$ \\
 & $0.8$                           & $0.254 \pm 0.022$ \\
\bottomrule
\end{tabular}
\caption{Hyperparameter sensitivity at $p=10$, $T=8$, $\sigma=0.5$, $\rho \sim \mathrm{Beta}(2,8)$. Bold marks the baseline configuration of Appendix~\ref{sec:ablation-setup}.}
\label{tab:hp_sensitivity}
\end{table}
Across all 25 off-baseline settings in Table~\ref{tab:hp_sensitivity}, main-effect F1 spans $0.254$ to $0.315$, against a baseline of F1 $= 0.276 \pm 0.013$. The largest deviations are $+0.039$ ($\tau_{\mathrm{start}} = 2.0$) and $-0.022$ (anneal fraction $0.8$). Both sit inside the seed-to-seed variation of the baseline itself, whose 5 seeds span $0.224$ to $0.362$ with a standard deviation of $0.044$. Three settings reach nominal significance against the baseline, all of them improvements, which is what 25 comparisons at the $5\%$ level produce by chance. No setting degrades performance. Network capacity is not a bottleneck across the ranges tested (hidden dim from 32 to 128, encoding dim from 16 to 64), and the logit-$\ell_2$ coefficient can be varied over two orders of magnitude ($0.001$ to $0.1$) without measurable impact. Doubling the training budget from 30{,}000 to 60{,}000 steps moves F1 by $+0.035$, again within the seed band, indicating that 30{,}000 steps is close to the training-time ceiling for this problem class.
\subsection{Method Ablations and Prior-Mismatch Robustness}
This subsection reports two families of ablations that keep the training recipe fixed while varying either the method or the deployment prior. The first (Table~\ref{tab:method_abl}) varies eleven off-baseline settings across seven design choices: encoder architecture, initialisation, reward structure, regularisation, mutual-information target, observation noise, and heredity assumption at evaluation, and quantifies each choice's contribution to the baseline F1 of $0.276 \pm 0.013$. The second (Table~\ref{tab:robustness}) evaluates the trained DABS policy on ground-truth data drawn from a mismatched prior, and compares its degradation to that of the static baseline.
\begin{table}[h!]
\centering
\small
\renewcommand{\arraystretch}{1.1}
\begin{tabular}{@{}llc@{}}
\toprule
\textbf{Component} & \textbf{Setting} & \textbf{Main F1} \\
\midrule
\multirow{2}{*}{Encoder}
 & \textbf{Sufficient statistics}       & $\mathbf{0.276 \pm 0.013}$ \\
 & Set encoder             & $0.326 \pm 0.008$ \\
\midrule
\multirow{2}{*}{Initialisation}
 & \textbf{Warm-start from static}       & $\mathbf{0.276 \pm 0.013}$ \\
 & Random                                & $0.306 \pm 0.034$ \\
\midrule
\multirow{2}{*}{Reward structure}
 & \textbf{Terminal sPCE}                & $\mathbf{0.276 \pm 0.013}$ \\
 & Dense per-step                        & $0.274 \pm 0.029$ \\
\midrule
\multirow{4}{*}{Regularisation}
 & \textbf{Logit-$\ell_2$ only}          & $\mathbf{0.276 \pm 0.013}$ \\
 & None                                  & $0.070 \pm 0.021$ \\
 & Cosine similarity only                & $0.117 \pm 0.020$ \\
 & Logit-$\ell_2$ + cosine               & $0.272 \pm 0.007$ \\
\midrule
\multirow{2}{*}{Training objective}
 & $\mathbf{I(z; h_T)}$ \textbf{(screening MI)} & $\mathbf{0.276 \pm 0.013}$ \\
 & $I(\theta; h_T)$ (total MI)           & $0.242 \pm 0.027$ \\
\midrule
\multirow{4}{*}{Noise level $\sigma$}
 & $0.25$                                & $0.271 \pm 0.016$ \\
 & $\mathbf{0.5}$                        & $\mathbf{0.276 \pm 0.013}$ \\
 & $1.0$                                 & $0.255 \pm 0.033$ \\
 & $2.0$                                 & $0.282 \pm 0.018$ \\
\midrule
\multirow{2}{*}{Heredity at evaluation}
 & \textbf{Strong (matched)}             & $\mathbf{0.276 \pm 0.013}$ \\
 & Weak ($z_{ij}$ unconstrained)         & $0.278 \pm 0.033$ \\
\bottomrule
\end{tabular}
\caption{Method ablations at $p=10$, $T=8$. Bold marks the baseline
configuration of Appendix~\ref{sec:ablation-setup}.}
\label{tab:method_abl}
\end{table}
The largest effect in Table~\ref{tab:method_abl} is regularisation. Removing the logit-$\ell_2$ penalty collapses F1 from $0.276$ to $0.070$, a $75\%$ drop. Without the penalty, the Gumbel-Softmax logits saturate early in training and the policy emits a near-constant design, confirming the collapse failure mode described in Section~3.3. Replacing the logit-$\ell_2$ penalty with a cosine-similarity penalty on consecutive designs does not prevent collapse (F1 $= 0.117$), and combining both recovers the baseline (F1 $= 0.272$). The logit-$\ell_2$ penalty is therefore the load-bearing regularisation choice.

The training-objective ablation is directionally consistent with the chain-rule prediction of Section~3.6. Targeting the total mutual information $I(\theta; h_T)$, as in DAD
\citep{ref1}, yields F1 $= 0.242$, a reduction of $0.034$ from the screening-specific target $I(z; h_T)$. Because $I(\theta; h_T) = I(z; h_T) + \mathbb{E}_z[I(\beta, \sigma^2; h_T \mid z)]$ and the conditional term admits no adaptive gain under the linear-Gaussian likelihood, resources spent estimating $\beta$ and $\sigma^2$ during training are wasted from the screening perspective. The reduction does not reach significance against the seed-to-seed variation at this problem class, so the ablation supports the prediction without establishing it.

The encoder ablation reverses the expectation we held when designing the method. Replacing the sufficient-statistics encoder with the set-equivariant encoder raises F1 from $0.276$ to $0.326$, a gain of $0.050$ with non-overlapping confidence intervals. At this problem class the per-pair encoder learns a more useful summary than the exact sufficient statistics do, despite the latter being sufficient in the formal sense. We attribute the difference to dimensionality. At $p = 10$ the statistic vector already carries roughly 1{,}600 entries into a width-32 projection, whereas the set encoder processes one pair at a time. Section~5.6 draws the consequence for scaling to larger $p$.

The remaining ablations do not change F1 much. Initialisation is a wash, with random initialisation at $0.306$ against $0.276$ for the warm start, a difference inside the seed band. Reward structure (terminal versus dense sPCE) differs by $0.002$, and noise level by at most $0.021$ across $\sigma \in \{0.25, 1.0, 2.0\}$. Under a heredity mismatch, training under strong heredity and evaluating under weak heredity, F1 does not drop at all ($0.278$ against $0.276$), showing that DABS is not brittle to a moderate violation of the training assumption.

\begin{table}[h!]
\centering
\small
\renewcommand{\arraystretch}{1.1}
\begin{tabular}{@{}llcc@{}}
\toprule
\textbf{Study} & \textbf{Evaluation prior} & \textbf{DABS F1} & \textbf{Static F1} \\
\midrule
\multirow{5}{*}{Sparsity mismatch}
 & $\pi = 0.1$                    & $0.200 \pm 0.019$ & $0.197$ \\
 & $\pi = 0.2$ (matched mean)     & $0.338 \pm 0.018$ & $0.281$ \\
 & $\pi = 0.3$                    & $0.371 \pm 0.013$ & $0.354$ \\
 & $\pi = 0.4$                    & $0.391 \pm 0.009$ & $0.339$ \\
 & $\pi = 0.5$                    & $0.420 \pm 0.016$ & $0.448$ \\
\midrule
\multirow{3}{*}{Beta hyperprior}
 & $\mathrm{Beta}(2, 8)$ (matched) & $0.272 \pm 0.011$ & $0.260$ \\
 & $\mathrm{Beta}(4, 6)$          & $0.393 \pm 0.009$ & $0.379$ \\
 & $\mathrm{Beta}(6, 4)$          & $0.425 \pm 0.009$ & $0.462$ \\
\bottomrule
\end{tabular}
\caption{Prior-mismatch robustness at $p=10$, $T=8$.}
\label{tab:robustness}
\label{tab:robustness}
\end{table}
Under sparsity mismatch, DABS leads the static baseline at $\pi = 0.2$, $0.3$ and $0.4$, by $0.057$, $0.017$ and $0.052$, and the two are level at $\pi = 0.1$ ($0.200$ against $0.197$). At $\pi = 0.5$ the ordering reverses and the static design leads by $0.028$, reaching F1 $= 0.448$ against $0.420$. The absolute F1 of both methods grows with $\pi$, because a higher active-factor base rate inflates F1 for any competent policy. The advantage of DABS is therefore concentrated in the sparse regime, and within that regime it does not require the sparsity of the training prior to match that of the evaluation.

The Beta hyperprior mismatch tells the same story. Under the matched Beta$(2, 8)$ prior DABS leads by $0.012$ and under Beta$(4, 6)$ by $0.014$. Under Beta$(6, 4)$, which biases the prior toward many-active configurations, the static design leads by $0.037$. As the evaluation prior puts more factors in the active set, F1 rises for both
methods. This is a base-rate effect rather than a real gain, because a higher
fraction of active factors inflates F1 for any competent policy. DABS holds its advantage across the sparse and moderately sparse mismatches. It loses that advantage only once the evaluation prior puts half or more of the factors in the active set, which is outside the screening regime the method targets.

\newpage
\section{Derivation of the joint distribution of the history} \label{apAc}
We derive the closed form of the integral introduced in
Eq.~\ref{eq:student-marginal},
\begin{equation}
p(h_T \mid z)
  \;=\;
  \iint
  p(h_T \mid z, \beta, \sigma^2)\,
  p(\beta \mid z)\,
  p(\sigma^2)\,
  d\beta\, d\sigma^2,
\end{equation}
where the parameter vector $\theta = (z, \beta, \sigma^2)$ is decomposed
into the discrete activity indicators $z$, the continuous effect sizes
$\beta$, and the noise variance $\sigma^2$. Conditioning on $z$, we
integrate out $(\beta, \sigma^2)$ analytically by exploiting the conjugate
normal--inverse-gamma structure of the prior.

\subsection*{Model specification}

Let $\mathbf{y} = (y_1,\ldots,y_T)^\top \in \mathbb{R}^T$ collect the observed responses in
history $h_T = \{(\xi_t, y_t)\}_{t=1}^T$, and let $X \in \mathbb{R}^{T \times d}$ be the
expanded design matrix with $d = p + \binom{p}{2}$ columns, comprising the main-effect
columns and the two-factor interaction columns. Given the activity pattern $z$, let
$\tau_z^2 \in \mathbb{R}^d$ denote the vector of prior variances, equal to
$\tau_{\text{active}}^2$ for active terms and $\tau_{\text{inactive}}^2$ otherwise. The
hierarchical model conditioned on $z$ is:
\begin{align}
\mathbf{y} \mid \beta, \sigma^2, z
  &\;\sim\; \mathcal{N}(X \beta, \sigma^2 I_T),
  \label{eq:app-lik} \\
\beta \mid \sigma^2, z
  &\;\sim\; \mathcal{N}(0, \sigma^2 \mathrm{diag}(\tau_z^2)),
  \label{eq:app-prior-beta} \\
\sigma^2
  &\;\sim\; \mathrm{IG}(\alpha_0,\; \beta_0),
  \label{eq:app-prior-sigma}
\end{align}
where $(\alpha_0, \beta_0)$ are the inverse-gamma hyperparameters
introduced in Section~3.1. This normal--inverse-gamma specification is the
conjugate prior for the Gaussian likelihood, enabling both integrals to be
computed in closed form.

\subsection*{Step 1: Integrating out \texorpdfstring{$\beta$}{beta\_z}}

Since the likelihood \eqref{eq:app-lik} and the prior
\eqref{eq:app-prior-beta} are both Gaussian conditionally on $\sigma^2$,
standard Gaussian identities give
\begin{equation}
\mathbf{y} \mid \sigma^2, z
  \;\sim\;
  \mathcal{N}\!\Big(0,\;
  \sigma^2 \underbrace{(I_T + X \mathrm{diag}(\tau_z^2) X^\top)}_{\displaystyle\Sigma_z}
  \Big),
\end{equation}
where $\Sigma_z := I_T + X \mathrm{diag}(\tau_z^2) X^\top$ encodes the covariance structure
induced by the activity pattern $z$ through the prior variances. Explicitly:
\begin{equation}
p(\mathbf{y} \mid \sigma^2, z)
  \;=\;
  (2\pi\sigma^2)^{-T/2}\,|\Sigma_z|^{-1/2}
  \exp\!\left(-\frac{\mathbf{y}^\top \Sigma_z^{-1} \mathbf{y}}{2\sigma^2}\right).
\label{eq:after-beta}
\end{equation}

\subsection*{Step 2: Integrating out \texorpdfstring{$\sigma^2$}{sigma\^{}2}}

We now integrate \eqref{eq:after-beta} against the
$\mathrm{IG}(\alpha_0, \beta_0)$ prior on $\sigma^2$. Collecting powers
of $\sigma^2$:
\begin{align}
p(\mathbf{y} \mid z)
  &= \int_0^\infty p(\mathbf{y} \mid \sigma^2, z)\,
     p(\sigma^2)\, d\sigma^2 \notag \\
  &= \frac{\beta_0^{\alpha_0}\,|\Sigma_z|^{-1/2}}
     {(2\pi)^{T/2}\,\Gamma(\alpha_0)}
     \int_0^\infty
     (\sigma^2)^{-(\alpha_0 + T/2 + 1)}
     \exp\!\left(-\frac{\beta_0 +
     \frac{1}{2}\mathbf{y}^\top \Sigma_z^{-1} \mathbf{y}}
     {\sigma^2}\right) d\sigma^2.
\end{align}
The integrand is an inverse-gamma kernel with updated parameters
\begin{equation}
\alpha_T \;:=\; \alpha_0 + \frac{T}{2},
\qquad
\beta_T \;:=\; \beta_0 + \frac{1}{2}\,\mathbf{y}^\top \Sigma_z^{-1} \mathbf{y},
\end{equation}
so the integral evaluates to $\Gamma(\alpha_T)/\beta_T^{\alpha_T}$,
yielding
\begin{equation}
p(\mathbf{y} \mid z)
  \;=\;
  \frac{\beta_0^{\alpha_0}\,\Gamma\!\left(\alpha_0 + \frac{T}{2}\right)}
       {(2\pi)^{T/2}\,\Gamma(\alpha_0)\,|\Sigma_z|^{1/2}}
  \left(\beta_0 + \frac{1}{2}\,\mathbf{y}^\top \Sigma_z^{-1}
  \mathbf{y}\right)^{-\left(\alpha_0 + \frac{T}{2}\right)}.
\label{eq:marginal-exact}
\end{equation}

\subsection*{Step 3: Identifying the Student-\texorpdfstring{$t$}{t} form}

Setting $\nu := 2\alpha_0$ (so the degrees of freedom are determined by
$\alpha_0$ alone) and defining the scale matrix
\begin{equation}
S_z := \frac{\beta_0}{\alpha_0}(I_T + X \mathrm{diag}(\tau_z^2) X^\top),
\label{eq:scale-final}
\end{equation}
expression \eqref{eq:marginal-exact} is exactly the density of the
multivariate Student-$t$ stated in Section~3.1:
\begin{equation}
p(h_T \mid z)
  \;=\;
  \frac{\Gamma\!\left(\frac{\nu + T}{2}\right)}
       {\Gamma\!\left(\frac{\nu}{2}\right)
        (\nu\pi)^{T/2}\,|S_z|^{1/2}}
  \left(1 + \frac{1}{\nu}\,\mathbf{y}^\top S_z^{-1}
  \mathbf{y}\right)^{-\frac{\nu + T}{2}},
\end{equation}
confirming that $h_T \mid z \sim \mathrm{St}_\nu(0, S_z)$. 

As stated in
Section~3.2, the degrees of freedom $\nu = 2\alpha_0$ and the scale matrix
$S_z$ are jointly determined by the design matrix induced by the history and the prior hyperparameters $(\alpha_0, \beta_0, \tau_{\text{active}}^2, \tau_{\text{inactive}}^2)$, with the activity pattern $z$ entering through the prior variance vector $\tau_z^2$.
\newpage
\section{Encoder details}
\label{apA}
The policy $\pi_\phi(\xi_t \mid h_{t-1})$ converts a variable-length history $h_{t-1} = \{(\xi_s, y_s)\}_{s=1}^{t-1}$ into per-factor design logits $\ell_t \in \mathbb{R}^p$, subject to two structural requirements: invariance to the order of past experiments and a fixed-dimensional output independent of $t$.

All encoders we consider share the same two-stage pipeline:
\begin{itemize}
    \item an encoder first summarises the history into a fixed-length vector $c_{t-1} \in \mathbb{R}^{d_{\mathrm{enc}}}$;
    \item an emitter MLP $\mu_\phi$ then maps this summary to the logits $\ell_t \in \mathbb{R}^p$. The emitter is a two-hidden-layer MLP with Softplus activations. At step $t = 0$, the encoder returns a learnable empty-history vector $c_0$ initialised to zero.
\end{itemize}
We describe both encoders used in this work. The
sufficient-statistics encoder is used in Case Studies 2 and 3 and throughout the
ablations of Appendix~\ref{apF}. The permutation-invariant set encoder is used
in Case Study 1. Table~\ref{tab:hparams} records the choice for each case
study.

\paragraph{Sufficient-statistics encoder.}
Let $d = p + \binom{p}{2}$ denote the number of main-effect plus two-way-interaction columns, let $X_{t-1} \in \mathbb{R}^{(t-1) \times d}$ stack the past designs expanded with those columns, and let $y_{t-1} \in \mathbb{R}^{t-1}$ collect the responses. Under the linear--Gaussian likelihood with conjugate Normal--Inverse-Gamma prior, the Student-$t$ posterior over $(\beta, \sigma^2)$ depends on the history only through the sufficient statistics $(X_{t-1}^\top X_{t-1},\ X_{t-1}^\top y_{t-1},\ y_{t-1}^\top y_{t-1})$. We assemble these statistics into the vector
\begin{equation}
    s_{t-1} = \left( \mathrm{vec}_\triangle\!\left(\frac{X_{t-1}^\top X_{t-1}}{t-1}\right),\;
\frac{X_{t-1}^\top y_{t-1}}{t-1},\; \frac{y_{t-1}^\top y_{t-1}}{t-1} \right)
\in \mathbb{R}^{\frac{d(d+1)}{2} + d + 1}
\end{equation}
where $\operatorname{vec}_{\triangle}(\cdot)$ extracts the upper-triangular entries of the symmetric matrix.
 
The normalisation by the number of observations in the history produces dimensionless inputs. For $\pm 1$ designs it maps diagonal entries of $X^\top X/(t-1)$ to 1. The statistic vector is then mapped to the encoding by a projection head
\begin{equation*}
\mu_{\mathrm{proj}} = \mathrm{Linear} \circ \mathrm{Softplus} \circ \mathrm{Linear} \circ \mathrm{LayerNorm}
\end{equation*}
with encoding width $d_{\mathrm{enc}}$.

Permutation invariance over past experiments holds by construction since each statistic is a symmetric function of the pairs $(\xi_s, y_s)$, and the output dimension is constant in $T$. The statistic dimension grows as $\mathcal{O}(d^2) = \mathcal{O}(p^4)$, which is tractable for the problem sizes considered here but not immediately scalable to very high-dimensional screening.

\paragraph{Set encoder.}
For comparison we also implement the set-equivariant encoder used by DAD~\citep{ref1}. Each pair $(\xi_s, y_s)$ is concatenated with sinusoidal positional features of the step index $s$ and mapped through a shared single-hidden-layer MLP $\psi_\phi$ to a per-step embedding. The embeddings are then summed:
\begin{equation}
e_s = \psi_\phi\big([\xi_s;\ y_s;\ \mathrm{PE}(s)]\big), \qquad c_{t-1} = \sum_{s=1}^{t-1} e_s.
\end{equation}
We use Softplus activation, a LayerNorm on the output to bound each contribution to a fixed magnitude, and sinusoidal positional features over $d_{\mathrm{pe}}$ frequency scales. Permutation invariance over the multiset holds because the summation is commutative. 

This encoder is fully general, as it does not assume sufficiency, and the
per-pair MLP learns to approximate the sufficient statistics from data.
Appendix~\ref{apF} finds that it outperforms the sufficient-statistics encoder
at the ablation problem class, by $0.050$ main-effect F1.
Each per-pair input has dimension $p + 1 + d_{\mathrm{pe}}$, so the encoder's
input dimension grows linearly in $p$ rather than as $\mathcal{O}(p^4)$.
\clearpage
\section{Pseudo-code of the training procedure}
\label{apAd}
Algorithm~\ref{alg:training} summarises the full training procedure.

\begin{algorithm}[h]
\caption{Training DABS}
\label{alg:training}
\begin{algorithmic}
\Require Prior $p(\theta)$, forward model $f$, factors $p$, budget $T$, contrastive samples $L$, batch size $B$, logit-$\ell_2$ weight $\mu$, temperature schedule $\tau_{\mathrm{start}} \rightarrow \tau_{\mathrm{end}}$, steps $S$, learning rate $\eta$
\Ensure Policy parameters $\phi$
\State Initialise policy $\pi_\phi$
\For{$s = 1, \dots, S$}
    \State Update temperature $\tau$ via annealing schedule
    \State \textit{Batched rollout:} Sample $\{(z^{(b)}, \beta^{(b)}, \sigma^{2(b)})\}_{b=1}^B \sim p(\theta)$. Initialise $h_0^{(b)} \gets \emptyset$ for $b = 1, \dots, B$.
    \For{$t = 1, \dots, T$}
        \For{$b = 1, \dots, B$}
            \State $\ell_t^{(b)} \gets \pi_\phi(h_{t-1}^{(b)})$
            \State $\xi_t^{(b)} \gets \mathrm{GumbelSoftmax}(\ell_t^{(b)}, \tau)$
            \State $y_t^{(b)} \sim \mathcal{N}(f(\xi_t^{(b)}, \beta^{(b)}), \sigma^{2(b)})$
            \State $h_t^{(b)} \gets h_{t-1}^{(b)} \cup \{(\xi_t^{(b)}, y_t^{(b)})\}$
        \EndFor
    \EndFor
    \State \textit{Contrastive sampling:} Sample $\{z_i^{(b)}\}_{i=1,\dots,L,\; b=1,\dots,B} \sim p(z)$
    \State \textit{Loss:} Compute $p(h_T^{(b)} \mid z_i^{(b)})$ for all $b, i$
    \State $\mathcal{L}(\phi) \gets \tfrac{1}{B} \sum_{b=1}^B \left[-\widehat{\mathrm{sPCE}}_L^{(b)}(\phi)\right] + \mu\, \mathcal{L}_{\mathrm{L2}}(\phi)$
    \State \textit{Update:} $\phi \gets \phi - \eta \nabla_\phi \mathcal{L}(\phi)$
\EndFor
\end{algorithmic}
\end{algorithm}
\newpage
\section{Pseudo-code of the full deployment and inference pipeline}
\label{apB}
Algorithm~\ref{alg:deployment} summarises the full deployment and inference pipeline.
\begin{algorithm}[!h]
\caption{Adaptive Experimentation and Posterior Inference}
\label{alg:deployment}
\begin{algorithmic}[1]
\Require Trained policy $\pi_\phi$, budget $T$, posterior draws $G$, burn-in $N_{\mathrm{burn}}$
\Ensure $P(z_k = 1 \mid h_T)$,
$P(z_{ij} = 1 \mid h_T)$, effect sizes $\hat{\beta}_k$ and $\hat{\beta}_{ij}$ with $95\%$ CI

\vspace{0.5em}
\State \textit{Phase 1: Run adaptive experiments}
\State $h_0 \gets \emptyset$
\For{$t = 1, \dots, T$}
    \State $\ell_t \gets \pi_\phi(h_{t-1})$; \quad $\xi_t \gets \mathrm{sign}(\ell_t)$
    \State Observe $y_t$ from experiment
    \State $h_t \gets h_{t-1} \cup \{(\xi_t, y_t)\}$
\EndFor

\vspace{0.5em}
\State \textit{Phase 2: Posterior inference}
\State Initialise $\rho^{(0)} \in (0,1)$
\State Initialise $z^{(0)} \sim \mathrm{Bernoulli}(\rho^{(0)})^p$
\State Initialise $z_{ij}^{(0)} \sim \mathrm{Bernoulli}(\pi_{\mathrm{int}}\, z_i^{(0)} z_j^{(0)})$ for all $i < j$

\For{$g = 1, \dots, G + N_{\mathrm{burn}}$}
    \State \textit{Step A (update $z$ jointly with its interactions)}:
    \For{$k = 1,\dots,p$}
        \If{$z_k = 1$}
            \State Propose $z_k' \gets 0$, and $z_{ij}' \gets 0$ for every pair touching $k$
        \Else
            \State Propose $z_k' \gets 1$, and $z_{ij}' \sim \mathrm{Bernoulli}(\pi_{\mathrm{int}})$ for every pair
            \State \quad touching $k$ whose other parent is active, $0$ otherwise
        \EndIf
        \State Accept $(z_k', z_{\mathrm{int}}')$ with probability $\alpha$ \Comment{Eq.~\ref{20}}
    \EndFor
    \State \textit{Step A$'$ (update $z_{\mathrm{int}}$)}: Sample $z_{ij}^{(g)} \mid \mathrm{rest}, h_T$
           for every pair with $z_i^{(g)} = z_j^{(g)} = 1$ \Comment{Eq.~\ref{23}}
    \If{$\rho$ has a $\mathrm{Beta}(a,b)$ hyperprior}
        \State \textit{Step B (update $\rho$)}: Sample $\rho^{(g)} \sim \mathrm{Beta}\big(a + \sum_k z_k^{(g)},\; b + p - \sum_k z_k^{(g)}\big)$
    \EndIf
    \State \textit{Step C (update $\sigma^2$)}: Sample $\sigma^{2(g)} \mid z^{(g)}, h_T$ \Comment{Eq.~\ref{22}}
    \State \textit{Step D (update $\beta$)}: Sample $\beta^{(g)} \mid z^{(g)}, \sigma^{2(g)}, h_T$ \Comment{Eq.~\ref{18}}
\EndFor

\State \Return $P(z_k = 1 \mid h_T)$, $P(z_{ij} = 1 \mid h_T)$, $\hat{\beta}_k$ and $\hat{\beta}_{ij}$ averaged over post-burn-in samples.
\end{algorithmic}
\end{algorithm}

Step A is a Metropolis--Hastings move rather than a Gibbs step, because strong
heredity makes the conditional for a single main-effect indicator degenerate
whenever that factor has an active interaction. Heredity therefore holds by
construction of the proposal rather than being restored afterwards.
Appendix~\ref{apC} derives the acceptance probability and validates the sampler
against exact enumeration at $p = 4$.
\newpage
\section{Posterior inference details}
\label{apC}
We provide details of the sampler used to approximate the posterior distribution $p(z, \beta \mid h_T)$ under the spike-and-slab prior and Gaussian likelihood. Let $X \in \mathbb{R}^{T \times d}$ denote the design matrix including main effects and interaction terms, where $d = p + \binom{p}{2}$, and let $y \in \mathbb{R}^T$ collect the observed responses. The likelihood is
\begin{equation}
y \mid \beta, \sigma^2 \sim \mathcal{N}(X \beta, \sigma^2 I).
\end{equation}

The prior over coefficients is given by a spike-and-slab distribution:
\begin{equation}
\beta_k \mid z_k, \sigma^2 \sim \mathcal{N}(0, \sigma^2\, \tau_{z_k}^2),
\qquad
z_k \sim \mathrm{Bernoulli}(\rho),
\end{equation}
where $\tau_{z_k}^2 = \tau_{\mathrm{active}}^2$ when $z_k = 1$ and $\tau_{z_k}^2 = \tau_{\mathrm{inactive}}^2$ when $z_k = 0$. Interaction indicators follow the strong-heredity constraint of Section~3.1, $z_{ij} \mid z_i, z_j \sim \mathrm{Bernoulli}(\pi_{\mathrm{int}}\, z_i z_j)$, which reduces to $z_{ij} = z_i z_j$ in the deterministic limit $\pi_{\mathrm{int}} = 1$. The prior variance vector $\tau_z^2$ concatenates the main-effect variances and the heredity-derived interaction variances.

Direct posterior computation is intractable due to the combinatorial structure
of $z \in \{0,1\}^p$. We therefore employ a Markov chain that alternates between updating $z$,
$z_{\mathrm{int}}$, $\rho$, $\sigma^2$ and $\beta$, with a
Metropolis--Hastings step for $z$ and Gibbs steps for the rest. The steps below are presented in the order the sweep executes them, which is
the order used in the code.

The primary inferential targets are the posterior activity probability for each factor $k$ and the corresponding effect-size estimate, computed from $G$ post-burn-in samples as
\begin{equation}
\hat{P}(z_k = 1 \mid h_T) = \frac{1}{G} \sum_{g=1}^{G} z_k^{(g)},
\qquad
\hat{\beta}_k = \frac{1}{G} \sum_{g=1}^{G} \beta_k^{(g)},
\end{equation}
where $z_k^{(g)} \in \{0, 1\}$ and $\beta_k^{(g)} \in \mathbb{R}$ are the samples at iteration $g$. A factor $k$ is declared active if
\begin{equation}
\hat{P}(z_k = 1 \mid h_T) > 0.5,
\end{equation}
and $95\%$ credible intervals are obtained from the empirical quantiles of $\{\beta_k^{(g)}\}_{g=1}^{G}$~\citep{ref31}.

\paragraph{Step A: updating $z_k$ jointly with its interactions.}
Under strong heredity, a state in which factor $k$ is inactive while one of its
child interactions is active has probability zero. Whenever factor $k$ has an
active child, the conditional $p(z_k \mid z_{-k}, h_T)$ therefore places all of
its mass on $z_k = 1$. A single-site Gibbs step always returns $z_k = 1$, so the
factor can never be switched off. We instead use
a Metropolis--Hastings move on $z_k$ together with the interaction indicators it
parents.

Let $C_k = \{(i,j) : i = k \text{ or } j = k\}$ be the pairs touching factor $k$,
and let $A_k \subseteq C_k$ be those whose other parent is currently active. If
$z_k = 1$, the proposal sets $z_k' = 0$ and $z_{ij}' = 0$ for every pair in
$C_k$. If $z_k = 0$, it sets $z_k' = 1$ and draws
$z_{ij}' \sim \mathrm{Bernoulli}(\pi_{\mathrm{int}})$ independently for each pair
in $A_k$, holding the remaining pairs at zero.

The proposal density for the children is exactly their prior under the proposed
state, so those terms cancel against the prior ratio. Explicitly,
turning $z_k$ off multiplies the prior ratio by
$\prod_{(i,j) \in A_k} \pi_{\mathrm{int}}^{-z_{ij}}(1-\pi_{\mathrm{int}})^{-(1-z_{ij})}$
from the interaction indicators of Eq.~\ref{eq9}, and the reverse proposal
contributes exactly its reciprocal. The two cancel, leaving only the
main-effect odds. The acceptance probability reduces to
\begin{equation}
\alpha = \min\!\left\{ 1,\;
\frac{p(h_T \mid z')}{p(h_T \mid z)}
\left(\frac{\rho}{1-\rho}\right)^{\!s} \right\},
\qquad
s = \begin{cases} +1, & z_k : 0 \to 1,\\[2pt] -1, & z_k : 1 \to 0,\end{cases}
\label{20}
\end{equation}
where $\rho$ is the current sample of the activity probability and the marginal likelihood is available in closed form:
\begin{equation}
\log p(h_T \mid z)
=
\log t_{2\alpha_0}\!\left(
y \,;\, 0,\;
\frac{\beta_0}{\alpha_0}\big(X\, \mathrm{diag}(\tau_z^2)\, X^\top + I\big)
\right).
\label{21}
\end{equation}
This closed-form Student-$t$ marginal is exact under the conjugate
normal--inverse-gamma prior, with
$\beta \mid \sigma^2, z \sim \mathcal{N}(0, \sigma^2\, \mathrm{diag}(\tau_z^2))$
from Eq.~\ref{eq8} and
$\sigma^2 \sim \mathrm{Inv\text{-}Gamma}(\alpha_0, \beta_0)$ from
Eq.~\ref{eq:model}, which we use throughout for data generation, training and
posterior inference.

The move costs two marginal likelihood evaluations per factor per sweep, the
same order as the Bernoulli conditional it replaces. When
$\pi_{\mathrm{int}} = 1$ the proposal is deterministic in both directions, the
interaction indicators are a function of their parents, and the move reduces to
a Metropolis step on a two-point space.

\paragraph{Step A$'$: updating the interaction indicators.}
Conditional on both parents being active, the conditional for $z_{ij}$ is not
degenerate. It is an ordinary Bernoulli draw,
\begin{equation}
P(z_{ij} = 1 \mid \mathrm{rest}, h_T)
=
\frac{p(h_T \mid z_{ij} = 1, \mathrm{rest})\, \pi_{\mathrm{int}}}
{p(h_T \mid z_{ij} = 1, \mathrm{rest})\, \pi_{\mathrm{int}}
+ p(h_T \mid z_{ij} = 0, \mathrm{rest})\, (1 - \pi_{\mathrm{int}})},
\label{23}
\end{equation}
with both marginal likelihoods given by Eq.~\ref{21}. Pairs whose parents are
not both active are held at zero by construction of the Step~A proposal, so no
separate heredity-enforcement step is required.

\paragraph{Validation.}
At $p = 4$ the admissible set contains 113 configurations, so the posterior can
be enumerated exactly. Across 60{,}000 draws, the Metropolis-Hastings sampler attains a total variation
distance of $0.003$ from the exact posterior. The Monte Carlo floor is $0.004$,
obtained by comparing the two halves of the chain against each other, so the
sampler is indistinguishable from the exact posterior at this chain length. The
single-site step that zeroes the children attains $0.176$ on the same problem.

\paragraph{Step B: sampling $\rho \mid z$.}
Under the $\mathrm{Beta}(a,b)$ hyperprior of Eq.~\ref{eq8}, $\rho$ has the
conjugate full conditional
$\rho \mid z \sim \mathrm{Beta}\big(a + \sum_{k=1}^{p} z_k,\;
b + p - \sum_{k=1}^{p} z_k\big)$, which is resampled once per sweep when that
hyperprior is used. The same $\mathrm{Beta}(a,b)$ prior is used during policy
training, where $\rho$ is drawn once per simulated instance. The initial value
of $\rho$ does not affect the stationary distribution and is discarded during
burn-in.

\paragraph{Step C: sampling $\sigma^2 \mid z, h_T$.}
Conditional on $z$ alone, with $\beta$ integrated out, the noise variance has an
inverse-gamma conditional posterior. Under the $\sigma$-scaled slab,
$y \mid \sigma^2, z \sim \mathcal{N}(0, \sigma^2 \Sigma_z)$, so
\begin{equation}
\sigma^2 \mid z, h_T \sim
\mathrm{InvGamma}\!\left(\alpha_0 + \tfrac{T}{2},\;
\beta_0 + \tfrac{1}{2}\, y^\top \Sigma_z^{-1} y\right),
\qquad
\Sigma_z = I_T + X\,\mathrm{diag}(\tau_z^2)\, X^\top ,
\label{22}
\end{equation}
as derived in Appendix~\ref{apAc}. Drawing $\sigma^2$ from this marginal
conditional rather than from a $\beta$-conditional form lets the sweep run $z$,
then $\sigma^2$, then $\beta$, so that each variable is drawn given the current
value of the others. Because $\rho$ depends only on $z$, its position among the last
three is immaterial. The $z$-update of Step A uses the Student-$t$ marginal of
Eq.~\ref{21}, which integrates $\beta$ and $\sigma^2$ out analytically. The
update therefore does not depend on the current values of $\beta$ and
$\sigma^2$.

\paragraph{Step D: sampling $\beta \mid z, \sigma^2, h_T$.}
Let $\tau_z^{-2}$ denote the prior precision vector under activity pattern $z$. Given the current $\sigma^2$ sample from Step C, and conditioned on $z$, the prior on $\beta$ is a diagonal Gaussian and the likelihood is linear--Gaussian. By Normal--Normal conjugacy the posterior is

\begin{equation}
\beta \mid z, \sigma^2, h_T \sim \mathcal{N}(\mu_{\mathrm{post}}, \Sigma_{\mathrm{post}}),
\quad
\Sigma_{\mathrm{post}} = \sigma^2 \left(X^\top X + \mathrm{diag}(\tau_z^{-2})\right)^{-1},
\quad
\mu_{\mathrm{post}} = \frac{1}{\sigma^2}\, \Sigma_{\mathrm{post}}\, X^\top y.
\label{18}
\end{equation}
\newpage
\section{Extended case study results}
This appendix reports the metric curves that did not appear in the main text for each of the three case studies of Section~5. In each subsection, the figures use the same DABS and baseline colour scheme as the corresponding main-text figures and share the same random-seed protocol (five seeds per $(p,T)$ cell, 95\% confidence intervals).
\subsection{Case Study 1: biotechnology benchmark (main effects, $T$ sweep at $p=15$)}
\label{app:5_1_extra}
\begin{figure}[!h]
\centering
\includegraphics[width=0.7\linewidth]{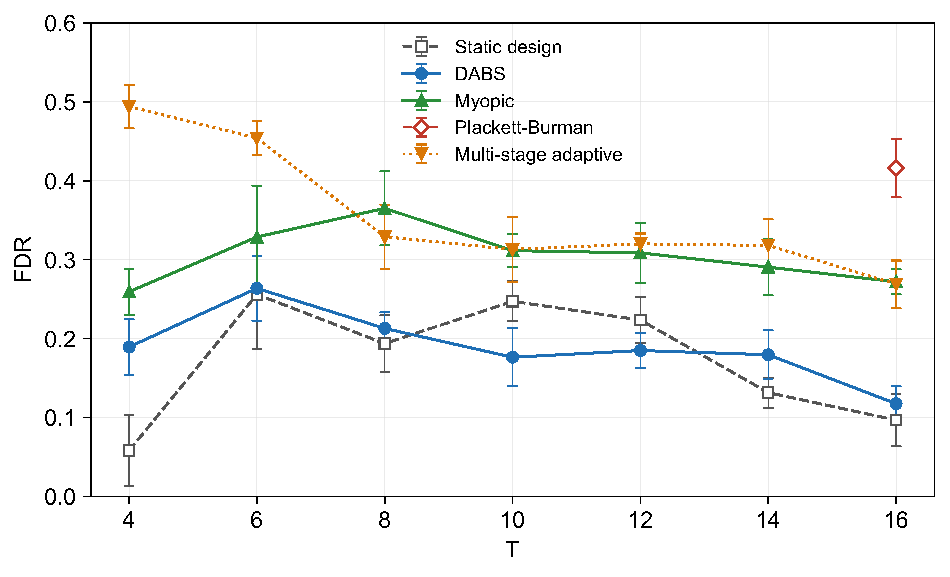}
\caption{Main-effect FDR versus budget $T$ at $p = 15$ for the biotechnology benchmark}
\label{fig:c1_FDR}
\end{figure}

Section~5.2 reports F1 and TPR versus the budget $T$ at $p = 15$. Figure~\ref{fig:c1_FDR} completes the picture with the corresponding FDR curves.

Figure~\ref{fig:c1_FDR} shows main-effect FDR versus the budget $T$.
The three Bayesian methods (DABS, the static design and the myopic baseline)
achieve substantially lower FDR than the two non-Bayesian baselines across the
sweep, staying between $0.06$ and $0.27$. Multi-stage adaptive maintains an FDR
between $0.36$ and $0.62$, and Plackett-Burman reaches $0.42$ at its native $T = 16$. Between the two Bayesian methods, the static design achieves lower FDR at the smallest budgets, but this reflects its very low recall in that regime (recall F1 $= 0.03$ at $T = 4$ from Section~5.2): the static design predicts almost no actives at all, so its false-positive rate is low by default. Above the identifiability boundary, DABS and the static design converge to comparable FDR values around $0.10$ at $T = 16$, consistent with the wedge prediction of Section~3.6.
\subsection{Case Study 2: HSC benchmark (main effects, $p$ sweep at $T = 16$)}
\label{app:case2_extra}
\begin{figure}[!h]
\centering
\includegraphics[width=0.7\linewidth]{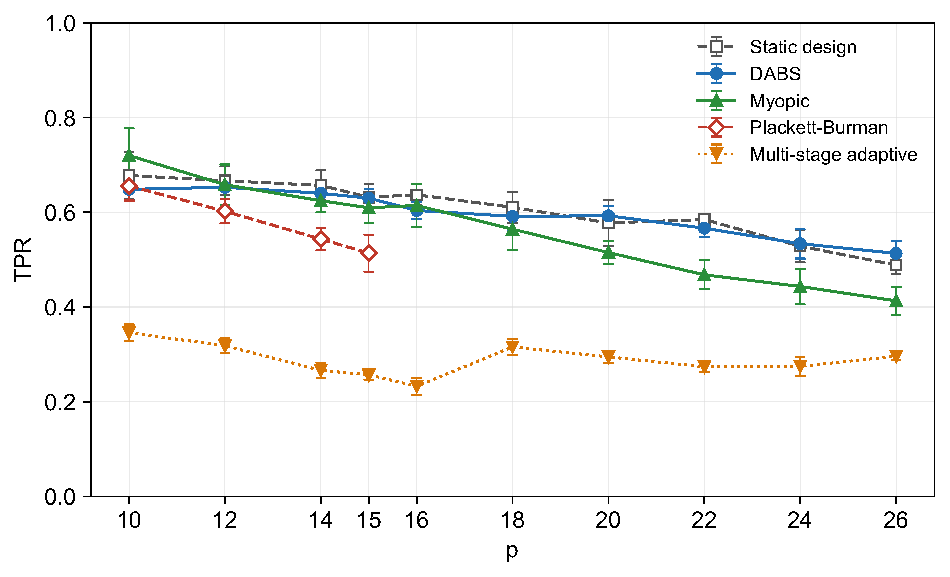}
\caption{Main-effect TPR versus number of candidate factors $p$ at $T = 16$ for the HSC benchmark. Bands are 95\% CIs over 5 seeds.}
\label{fig:c2_TPR}
\end{figure}

Section~5.3 reports F1 and FDR versus $p$ at fixed $T = 16$. Figure~\ref{fig:c2_TPR} completes the picture with the true positive rate for the same sweep.
TPR declines for every method as $p$ grows, as the fixed budget $T = 16$ is spread over an increasing number of candidate factors. In the identifiable regime $p \leq 15$, all methods except multi-stage adaptive achieve TPR between 0.5 and 0.7, with Plackett-Burman competitive up to its native $p = 15$. From $p = 18$ onward, DABS and the static design track each other closely (TPR
between $0.42$ and $0.52$). DABS stays above the myopic baseline at every $p$ in
this range, while the static design falls slightly below it at $p = 18$ and
$p = 20$ before overtaking it again. The myopic baseline degrades fastest with
$p$, reaching $0.41$ at $p = 26$. Multi-stage adaptive stays between $0.24$ and
$0.37$ throughout, well below the other methods. Its multi-stage pruning discards candidates aggressively at each stage, so recall on the active set remains low regardless of $p$.
\subsection{Case Study 3: metallurgical benchmark (main effects + interactions, $T$ sweep at $p \in \{4, 6, 8\}$)}
\label{app:case3_extra}
\begin{figure}[!h]
\centering
\includegraphics[width=\linewidth]{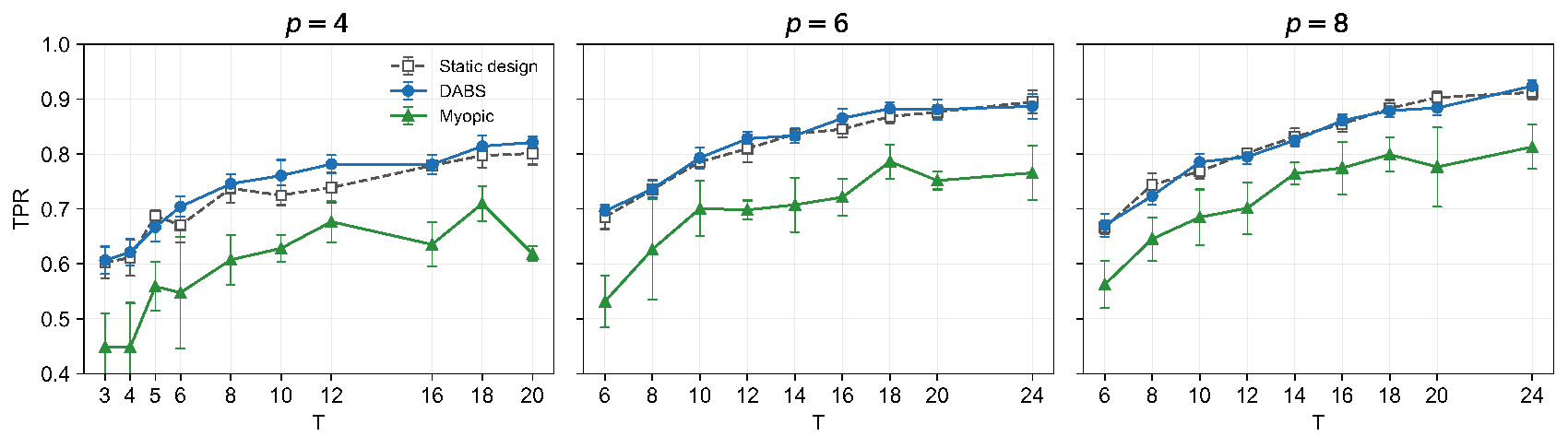}
\caption{Main-effect TPR versus budget $T$ across $p \in \{4, 6, 8\}$}
\label{fig:c3_TPR_main}
\end{figure}

\begin{figure}[!h]
\centering
\includegraphics[width=\linewidth]{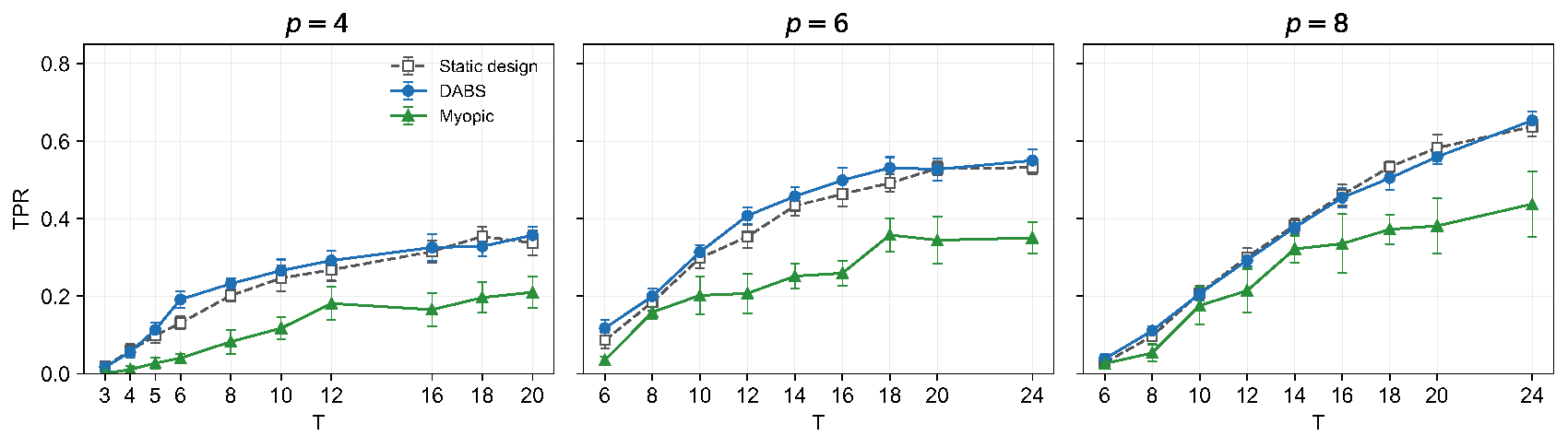}
\caption{Two-factor interaction TPR versus budget $T$ across $p \in \{4, 6, 8\}$}
\label{fig:c3_TPR_int}
\end{figure}

\begin{figure}[!h]
\centering
\includegraphics[width=\linewidth]{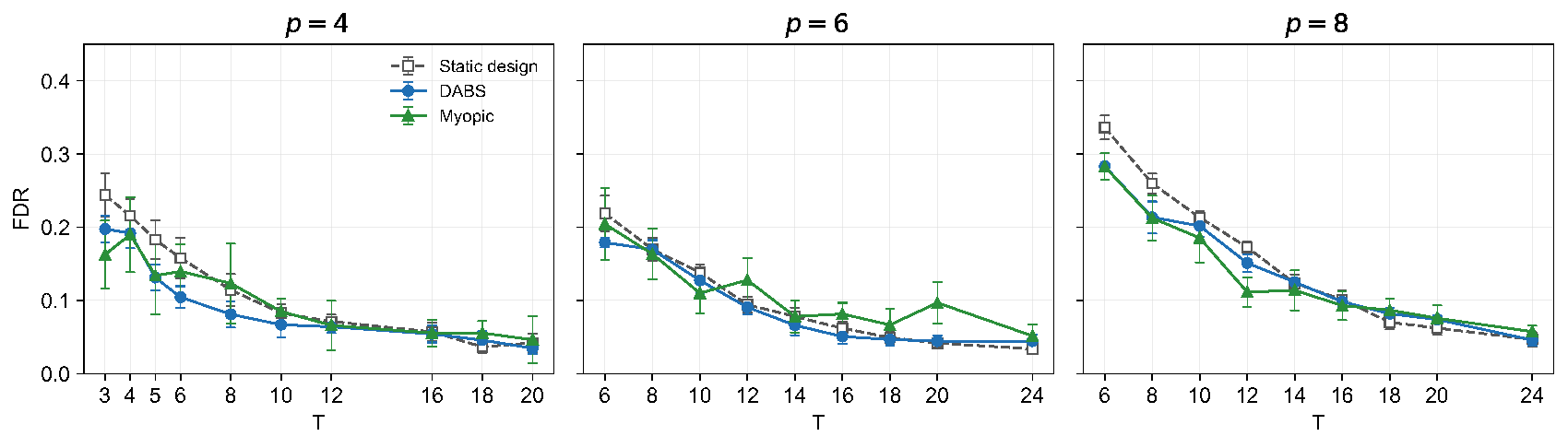}
\caption{Main-effect FDR versus budget $T$ across $p \in \{4, 6, 8\}$}
\label{fig:c3_FDR_main}
\end{figure}

\begin{figure}[!h]
\centering
\includegraphics[width=\linewidth]{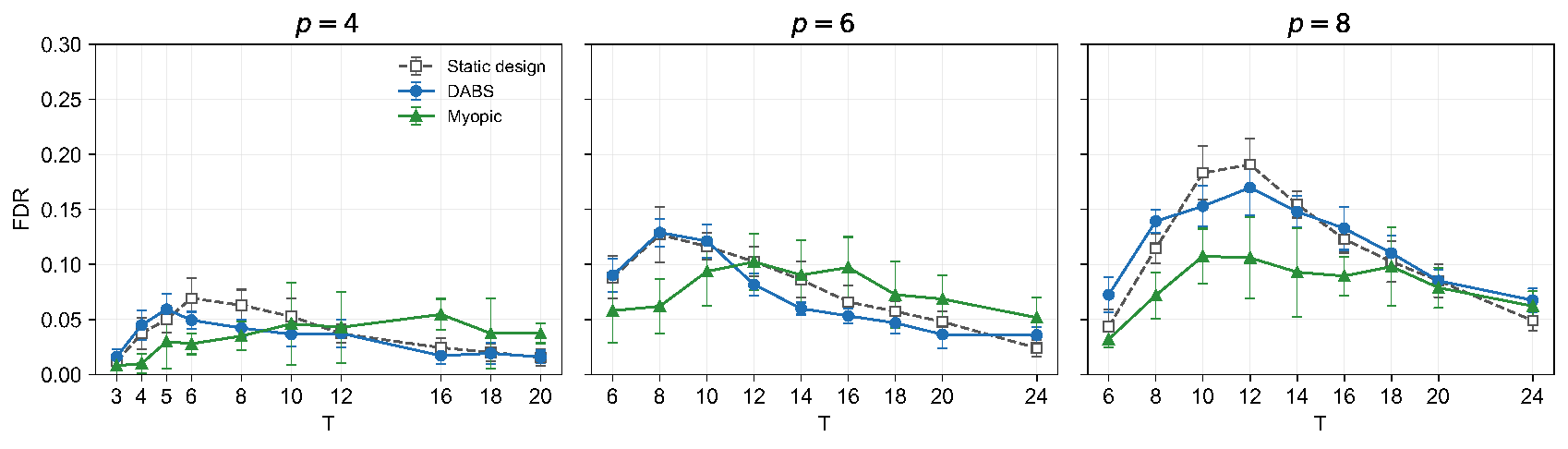}
\caption{Two-factor interaction FDR versus budget $T$ across $p \in \{4, 6, 8\}$}
\label{fig:c3_FDR_int}
\end{figure}

Section~5.4 reports F1 for main effects and two-factor interactions across the three problem sizes. Figures~\ref{fig:c3_TPR_main}--\ref{fig:c3_FDR_int} complete the picture with the corresponding TPR and FDR curves.

Main-effect TPR and FDR behave as ceiling metrics for the two amortised policies, mirroring the F1 story of Section~5.4: DABS and the static design track each other closely, with higher TPR and lower
FDR than the myopic baseline.
The interaction TPR and FDR curves show a clearer three-way separation at $p = 6$ in the mid-budget window, again matching the F1 pattern.
At $p = 8$ the three methods are much closer. The myopic baseline matches or
exceeds DABS on interaction TPR at $T = 10$ and $T = 14$ and attains the lowest
interaction FDR at those budgets, so no method holds a consistent advantage across
the sweep. All three converge at the high-budget end as the interaction problem
approaches its own identifiability ceiling.
\newpage
\section{Credible interval calibration}
\label{apI}
The sampler of Appendix~\ref{apC} returns posterior draws over the effect sizes, and credible intervals are obtained from their empirical quantiles. This appendix evaluates whether those intervals are calibrated, meaning that an interval stated at level $\alpha$ contains the true coefficient a fraction $\alpha$ of the time.

We reuse the trained policies of Section~5 without modification. For each checkpoint we run 100 rollouts. In every rollout the ground truth $(z, \beta, \sigma^2)$ is drawn from the generative model, the policy selects the $T$ designs, and the sampler is run with the settings used throughout Section~5, namely 200 draws after 100 burn-in steps. For each coefficient we form the central interval at level $\alpha$ from the empirical quantiles of the post-burn-in draws and record whether the true coefficient falls inside it. Results are pooled over the full sweep of each case study and over five training seeds, giving between roughly 10,000 and 110,000 intervals per group.

Coverage is reported separately for active and inactive coefficients. The two are not comparable, because an inactive coefficient is drawn from the spike with $\tau_{\text{inactive}} = 0.01$ and its interval is correspondingly narrow. Active coefficients are further split by whether the sampler recovered them, meaning $\hat{P}(z_k = 1 \mid h_T) > 0.5$. This separates the calibration of the posterior from the accuracy of the screening decision. Coverage on a coefficient that has been wrongly declared inactive measures the second, not the first.
\begin{table}[!h]
\centering
\begin{tabular}{lcccc}
\toprule
& \multicolumn{4}{c}{Nominal level} \\
\cmidrule(lr){2-5}
Coefficient group & $0.50$ & $0.80$ & $0.90$ & $0.95$ \\
\midrule
\multicolumn{5}{l}{\textit{Case Study 1} ($p = 15$, $T$ from 4 to 16, $\mathrm{SNR} = 4$)} \\
\quad Main, inactive             & 0.556 & 0.861 & 0.945 & 0.978 \\
\quad Main, active and recovered & 0.478 & 0.774 & 0.874 & 0.926 \\
\quad Main, all active           & 0.242 & 0.504 & 0.658 & 0.775 \\
\midrule
\multicolumn{5}{l}{\textit{Case Study 2} ($T = 16$, $p$ from 10 to 26, $\mathrm{SNR} = 1$)} \\
\quad Main, inactive             & 0.664 & 0.942 & 0.987 & 0.997 \\
\quad Main, active and recovered & 0.380 & 0.716 & 0.849 & 0.916 \\
\quad Main, all active           & 0.234 & 0.561 & 0.743 & 0.850 \\
\midrule
\multicolumn{5}{l}{\textit{Case Study 3} ($p = 8$, $T$ from 6 to 24, $\mathrm{SNR} = 4$)} \\
\quad Main, inactive             & 0.579 & 0.880 & 0.955 & 0.982 \\
\quad Main, active and recovered & 0.483 & 0.785 & 0.883 & 0.934 \\
\quad Main, all active           & 0.404 & 0.702 & 0.823 & 0.893 \\
\quad Interaction, inactive      & 0.530 & 0.835 & 0.928 & 0.968 \\
\quad Interaction, active        & 0.238 & 0.477 & 0.632 & 0.753 \\
\bottomrule
\end{tabular}
\caption{Empirical coverage against nominal level, pooled over each sweep and five seeds.}
\label{tab:calibration}
\end{table}

Table~\ref{tab:calibration} reports the calibration curve. At the 95\% level the
main-effect intervals sit slightly below nominal in all three case studies once
support recovery is correct, reaching $0.926$, $0.916$ and $0.934$. Inactive
coefficients are conservative throughout, covering more often than the nominal
level at every level tested. This is expected, since the posterior concentrates
on the same spike from which the true inactive coefficients are drawn.

At tighter levels the signal-to-noise ratio separates the case studies.
Coverage conditional on recovery at the 50\% level is $0.483$ in Case Study 3
and $0.478$ in Case Study 1, which share a signal-to-noise ratio of 4, against
$0.380$ in Case Study 2, where it is 1. The two matched case studies land
within $0.005$ of each other despite differing in dimensionality and budget,
with $p = 15$ against $p = 8$. When the signal is weak the spike-and-slab prior
shrinks the posterior toward zero, so the central half of the posterior can sit
between zero and the true value, and a narrow interval misses more often than
nominal. Wider intervals reach far enough to absorb the shrinkage, which is why
the three case studies converge again at the 95\% level.
\begin{table}[!h]
\centering
\begin{tabular}{lccccc}
\toprule
& \multicolumn{3}{c}{Coverage at $95\%$} & \multicolumn{2}{c}{Mean width} \\
\cmidrule(lr){2-4} \cmidrule(lr){5-6}
Setting & Inactive & Active, recovered & Active, missed & Inactive & Active \\
\midrule
\multicolumn{6}{l}{\textit{Case Study 1}, budget sweep at $p = 15$} \\
\quad $T = 4$   & 0.995 & 0.863 & 0.684 & 1.667 & 2.604 \\
\quad $T = 8$   & 0.988 & 0.925 & 0.652 & 1.159 & 2.466 \\
\quad $T = 12$  & 0.966 & 0.927 & 0.622 & 0.574 & 2.098 \\
\quad $T = 16$  & 0.960 & 0.929 & 0.618 & 0.518 & 2.195 \\
\midrule
\multicolumn{6}{l}{\textit{Case Study 2}, dimensionality sweep at $T = 16$} \\
\quad $p = 10$  & 0.990 & 0.932 & 0.713 & 0.564 & 1.149 \\
\quad $p = 16$  & 0.995 & 0.915 & 0.772 & 0.933 & 1.568 \\
\quad $p = 20$  & 0.999 & 0.906 & 0.766 & 1.192 & 1.840 \\
\quad $p = 26$  & 0.999 & 0.905 & 0.776 & 1.408 & 1.937 \\
\midrule
\multicolumn{6}{l}{\textit{Case Study 3}, budget sweep at $p = 8$} \\
\quad $T = 6$   & 1.000 & 0.918 & 0.781 & 1.185 & 1.627 \\
\quad $T = 12$  & 0.984 & 0.934 & 0.717 & 0.505 & 1.010 \\
\quad $T = 18$  & 0.974 & 0.942 & 0.566 & 0.232 & 0.733 \\
\quad $T = 24$  & 0.962 & 0.930 & 0.654 & 0.106 & 0.447 \\
\bottomrule
\end{tabular}
\caption{Coverage and mean width of $95\%$ credible intervals for main effects, pooled over five seeds.}
\label{tab:coverage_budget}
\end{table}
Table~\ref{tab:coverage_budget} shows how coverage and width behave along each sweep.
Coverage conditional on recovery is stable, between $0.86$ and $0.94$ across
every setting tested. In Case Study 1 it rises from $0.863$ at $T = 4$ to
$0.929$ at $T = 16$, and in Case Study 2 it drifts from $0.932$ at $p = 10$ to
$0.905$ at $p = 26$ while the mean active width grows from $1.15$ to $1.94$, so
the posterior answers a harder problem by widening and gives up a little
calibration as it does. Mean active width contracts from $2.60$ to about $2.20$
as the budget grows in Case Study 1 and from $1.63$ to $0.45$ in Case Study 3.
Coverage on missed factors is lower throughout but does not collapse at the
95\% level, because a factor with inclusion probability below $0.5$ is still
active in a substantial fraction of draws and its interval retains a wide tail.

Interaction intervals are the weakest case. In Case Study 3 coverage on active
interactions rises from $0.63$ at $T = 6$ to $0.85$ at $T = 24$, below the
nominal level throughout. These per-budget values are not tabulated.
Table~\ref{tab:calibration} reports the calibration curve pooled over the sweep, which
sits well under the diagonal.

This mirrors the interaction recovery results of Section~5.4, where interaction F1 reaches approximately $0.70$ at $p = 8$ and $T = 24$. Under strong heredity an interaction can only be identified once both parents have been resolved, so its posterior remains diffuse for longer than that of a main effect.

These results support a specific claim. The $95\%$ credible intervals produced at deployment achieve coverage slightly below nominal, between $0.92$ and $0.93$, for main
effects whose activity is correctly identified, across all three case studies, and are conservative for coefficients correctly ruled out. Coverage on the remaining coefficients is governed by the accuracy of the screening decision rather than by the posterior itself, so intervals inherit the quality of the support recovery reported in Section~5. Narrower intervals, and interaction intervals at tight budgets, should be treated as indicative rather than calibrated.
\newpage
\section{On the use of implicit DAD (iDAD)} \label{apD}
Recent work has proposed implicit Deep Adaptive Design (iDAD)~\citep{ref6}, which replaces explicit likelihood evaluation in the expected information gain objective with a learned critic network and enables amortized design optimization in simulator-based settings with intractable likelihoods.

In our setting, however, the model admits a conjugate Normal--Inverse-Gamma prior with a linear Gaussian likelihood, yielding a closed-form Student-$t$ marginal likelihood. This quantity can therefore be evaluated exactly within the prior-contrastive estimator.

As a result, adopting iDAD would replace an exact computation with a learned approximation, introducing additional estimator variance and potential bias without a compensating benefit. We therefore retain an explicit likelihood formulation throughout.
\newpage
\section{Result details and discussion}
\label{apE}
This appendix collects the exact hyperparameters, architectures, and evaluation settings used across the three case studies of Section~5.
Case Studies 2 and 3 use the sufficient-statistics encoder of
Appendix~\ref{apA} with a two-hidden-layer MLP emitter, and Case Study 1 uses
the set encoder, as recorded in Table~\ref{tab:hparams}. The encoder belongs to
the DABS policy alone and does not enter the static or myopic baselines, so this
choice understates DABS in Case Studies 2 and 3 rather than favouring it. All training uses stochastic gradient descent on the combined objective $\mathcal{L}(\phi) = -\widehat{\mathrm{sPCE}}_L(\phi) + \mu\, \mathcal{L}_{\mathrm{L2}}(\phi)$ defined in Section~3.2.

Table~\ref{tab:hparams} summarises the model priors, training schedule, and evaluation protocol for each case study.

\begin{table}[ht]
\centering
\small
\begin{tabular}{lccc}
\toprule
                                   & Case Study 1 & Case Study 2 & Case Study 3 \\
                                   & (biotech, §5.2) & (HSC, §5.3) & (GH4738, §5.4) \\
\midrule
\textbf{Model priors} \\
\quad Number of factors $p$        & $15$ & $\{10, 12, \ldots, 26\}$ & $\{4, 6, 8\}$ \\
\quad Experimental budget $T$      & $\{4, 6, \ldots, 16\}$ & $16$ & sweep, see §5.4 \\
\quad Sparsity prior $\rho$        & $0.2$ (fixed) & $\rho \sim \mathrm{Beta}(4, 6)$ & $0.5$ (fixed) \\
\quad Active slab std $\tau_{\mathrm{active}}$ & $2.0$ & $1.0$ & $2.0$ \\
\quad Inactive slab std $\tau_{\mathrm{inactive}}$ & $0.01$ & $0.01$ & $0.01$ \\
\quad Nominal noise std $\sigma$ (sets $\beta_0$) & $0.5$ & $0.5$ & $0.25$ \\
\quad InvGamma shape $\alpha_0$    & $3.0$ & $3.0$ & $3.0$ \\
\quad InvGamma scale $\beta_0$     & $1.0$ & $1.0$ & $0.25$ \\
\quad Interaction prior $\pi_{\mathrm{int}}$ & $0.5$ & $0.5$ & $0.5$ \\
\quad Strong heredity              & yes & yes & yes \\
\midrule
\textbf{DABS training} \\
\quad Optimisation steps $S$       & $30{,}000$ & $30{,}000$ & $30{,}000$ \\
\quad Learning rate                & $10^{-3}$ & $10^{-3}$ & $10^{-3}$ \\
\quad Adam $\beta_1, \beta_2$      & $0.9, 0.999$ & $0.9, 0.999$ & $0.9, 0.999$ \\
\quad Outer samples (rollout batch)& $6000$ & $2000$--$6000$ & $6000$ \\
\quad Contrastive samples $L$      & $512$ & $256$--$512$ & $512$ \\
\quad Logit-$\ell_2$ weight $\mu$  & $0.01$ & $0.01$ & $0.01$ \\
Gumbel temperature schedule & \multicolumn{3}{c}{$\tau_{\mathrm{start}} = 1.0 \to \tau_{\mathrm{end}} = 0.5$ over $80\%$ of training} \\
\midrule
\textbf{Static sPCE baseline} \\
\quad Optimisation steps           & $2000$ & $2000$ & $2000$ \\
\quad Learning rate                & $5 \times 10^{-3}$ & $5 \times 10^{-3}$ & $5 \times 10^{-3}$ \\
\quad Contrastive samples $L$      & $6000$ & $1500$--$6000$ & $6000$ \\
\midrule
\textbf{Architecture} \\
\quad Encoder                      & set enc. & suff. stats & suff. stats \\
\quad Encoding dimension $d_{\mathrm{enc}}$ & $32$ & $32$ & $32$ \\
\quad MLP hidden dimension         & $64$ & $64$ & $64$ \\
\quad Hidden layers (emitter)      & $2$ & $2$ & $2$ \\
\quad Activation                   & Softplus & Softplus & Softplus \\
\midrule
\textbf{Evaluation} \\
\quad Rollouts per policy          & $100$ & $100$ & $100$ \\
\quad Training seeds               & $5$ & $5$ & $5$ \\
\quad Gibbs samples                & $200$ & $200$ & $200$ \\
\quad Gibbs burn-in                & $100$ & $100$ & $100$ \\
\quad Classification threshold     & $0.5$ & $0.5$ & $0.5$ \\
\bottomrule
\end{tabular}
\caption{Hyperparameters and evaluation protocol for the three case studies.
The noise standard deviation $\sigma$ is nominal and sets the
inverse-gamma scale through $\beta_0 = \sigma^2(\alpha_0 + 1)$, so that
$\sigma^2$ is the prior mode.}
\label{tab:hparams}
\end{table}

The Plackett-Burman baseline uses the native run counts
$T \in \{8, 12, 16, 20, 24\}$ for its 8-, 12-, 16-, 20- and 24-run
constructions. The myopic baseline shares the same Bayesian model and Gibbs decoder as DABS. At each step it evaluates the one-step expected information gain over a candidate set of enumerated (small $p$) or randomly sampled ($p \geq 10$) designs and selects the maximiser. The multi-stage adaptive baseline runs $K = 3$ stages with keep-fraction $0.5$, using ridge-regularised linear estimates of marginal effects to prune inactive candidates between stages.

All experiments were run on NVIDIA H100 80GB GPUs on a university HPC cluster. A typical DABS training run takes approximately 10 minutes at $p = 8$, $T = 8$, and roughly 45 minutes at $p = 20$, $T = 16$. Evaluation with the Gibbs sampler takes approximately 5 minutes per checkpoint on CPU. Each reported F1, TPR, and FDR value is the mean over 100 rollouts computed independently for each of the 5 training seeds.
\clearpage
\newpage
\section{Visualisation of adaptive behaviour} \label{apK}
\begin{figure}[!h]
    \centering
    \includegraphics[width=\linewidth]{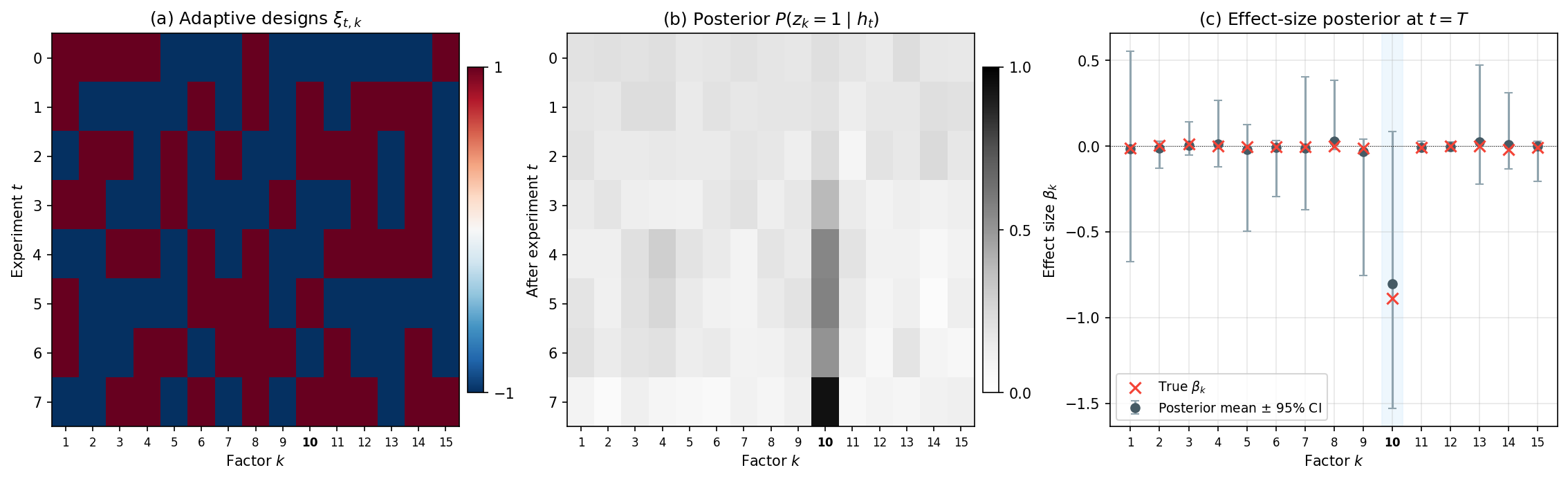}
    \caption{Illustration of adaptive behavior in a single rollout}
    \label{dd}
\end{figure}

To illustrate the behaviour of the learned policy, Figure \ref{dd} shows a single rollout of the sequential design process. Panel (a) displays the sequence of selected designs $\xi_{t,k} \in \{-1, +1\}$, where rows correspond to experiments and columns to factors. The early designs exhibit substantial variability across factors, reflecting an exploratory phase. As data are collected, the designs become more structured, with repeated interventions on a subset of factors, indicating a shift toward exploiting informative directions.

Panel (b) shows the evolution of the posterior inclusion probabilities $P(z_k = 1|h_t)$. Starting from a diffuse prior, the posterior progressively concentrates, with one factor emerging as clearly active while the others are shrunk toward low inclusion probabilities.

Panel (c) presents the posterior distribution of the effect sizes at the final step $t=T$, including posterior means and 95\% credible intervals. The active factor is accurately identified, while inactive effects are effectively shrunk toward zero.

Together, these panels illustrate how the policy adapts over time, transitioning from broad exploration to targeted experimentation, and enabling reliable recovery of the sparse active set.

\end{document}